# A neural operator for predicting vibration frequency response curves from limited data

D. Bluedorn[1], A. Badawy[2], B. E. Saunders[3], D. Roettgen[3], and A. Abdelkefi[1*]

[1]Department of Mechanical and Aerospace Engineering, New Mexico State University, Las Cruces, NM 88003, USA.

[2]Knight Foundation School of Computing and Information Science, Florida International University, Miami, FL 33199, USA.

[3]Sandia National Laboratories, Albuquerque, NM 87123, USA.

*Email: abdu@nmsu.edu

**Abstract**

In the design of engineered components, rigorous vibration testing is essential for performance validation and identification of resonant frequencies and amplitudes encountered during operation. Performing this evaluation numerically via machine learning has great potential to accelerate design iteration and make testing workflows more efficient. However, dynamical systems are conventionally difficult to solve via machine learning methods without using physics-based regularizing loss functions. To properly perform this forecasting task, a structure that has an inspectable physical obedience can be devised without the use of regularizing terms from first principles. The method employed in this work is a neural operator integrated with an implicit numerical scheme. This architecture enables operators to learn of the underlying state-space dynamics from limited data, allowing generalization to untested driving frequencies and initial conditions. This network can infer the system's global frequency response by training on a small set of input conditions. As a foundational proof of concept, this investigation verifies the machine learning algorithm with a linear, single-degree-of-freedom system, demonstrating implicit obedience of dynamics. This approach demonstrates 99.87% accuracy in predicting the Frequency Response Curve (FRC), forecasting the frequency and amplitude of linear resonance training on 7% of the bandwidth of the solution. By training machine learning models to internalize physics information rather than trajectory, better generalization accuracy can be realized, vastly improving the timeframe for vibration studies on engineered components.

**Keywords:** Dynamical systems, operator networks, data science, machine learning, modeling and simulation

## 1. Introduction

Vibration analysis is a critical consideration in the design of mechanical systems. Most components in aerospace and automotive applications can suffer from operational wear caused by excessive or unexpected vibration, which can lead to expensive and possibly dangerous system failures. The study of failure analysis is awash with examples of how over-simplified models of complex systems can lead to unforeseen failures. One prominent example of this is the wing failures of the Lockheed L-188

Electra due to whirling mode wing flutter in both Braniff Airways Flight 542 [1] and Northwest Airlines Flight 710 [2]. The self-sustaining gyroscopically-initiated flutter caused by the aircraft's wing-mounted propellers caused resonance to appear in turbulence at high airspeeds. These failures motivated the discovery that, despite thousands of hours of testing during the design of the aircraft, on whirl flutter in particular, the flight conditions encountered caused more deflection of the thrust axis than was expected, accentuating the flutter. It is estimated that the investigation of this failure cost Lockheed 2.5 million dollars, and later retrofits to solve it cost $800,000 per aircraft [3, 4]. These failures highlighted that even extensive testing can overlook critical resonance phenomena when analytical models underestimate the coupling between structural modes.

The study of component and system vibration is an expense to the designer, but far less costly than failure analysis and amelioration. It is for this reason that the study of component dynamics is carefully performed to predict vibration amplitudes and ensure reliable operation of mechanical systems. This analysis is done by modeling the component dynamics either theoretically from physical principles or using a data-driven method via experimentation [5, 6]. This process, however, is complicated by the variety and complexity of dynamical models, which may have interacting timescales causing stiffness [7, 8], unique resonances and nonlinearities [9-11], and other difficulties. As indicated by the vast space of nonlinear models, empirical workflows, and numerical models available, no single framework is fully sufficient to capture all important effects without a significant body of ground-truth experimental data. Additionally, the dataset sizes involved are often cumbersome for real-world systems with multiple degrees of freedom at high frequencies [12].

An essential part of engineering design is the experimental characterization of a system's frequency properties using harmonically driven vibration testing. This allows the resonances of the system to be known and knowledgeably considered during design iteration, and many workflows exist to allow testing on systems with unique characteristics [13-15]. This methodology can be time-consuming, as every frequency and condition must be thoroughly tested to reach a steady state. This can be accelerated by the use of swept sine testing, often at the expense of accuracy in the frequency response curve, especially if the data-processing technique is not well-suited to the system of study [16]. Finite Element Analysis (FEA) and similar computational modeling techniques are very helpful accelerators, but they often suffer from inaccuracies caused by assumptions made about boundary conditions or simplified material models [17, 18].

Dynamical system modeling is a high-profile problem in machine learning, and recent advances have begun to address this challenge by enabling data-driven representations of dynamic behavior [19-21]. Neural networks, recurrent architectures, and physics-informed frameworks can approximate system operators directly from observed data, learning input-output relationships that would otherwise require explicit modeling. However, most conventional networks treat dynamics by direct encoding of the limited data in a regression task, which can limit the extrapolation ability of the model. This motivates the use of operator learning approaches that encode the governing relationships themselves, rather than the particular solution, enabling far more robust prediction of frequency-domain behavior from sparse time-domain data. Experimental implementations of these methodologies use vanilla neural networks, but several are branching out into more generalizable neural state space models [22, 23]. These methods are more capable of extrapolation to untrained trajectories, especially when the dataset does not encompass a significant portion of the target response.

From the existing body of literature, it is evident that most machine learning investigations in vibration dynamics fall into two main categories. There are studies focused on simple numerical models like low-dimensional oscillators or idealized state-space systems where algorithms achieve remarkable accuracy in reproducing time histories from minimal data [7, 22, 24, 25]. On the other side of the spectrum, some studies applied neural networks directly to experimental data from real components, using sensor responses to identify qualitative behaviors, such as crack formation or joint conditions [20, 21, 23, 26, 27]. However, these two methodologies lack a unification of purpose, and high-fidelity machine learning extrapolation remains a challenge. Methods developed for simple systems do not extend easily to experimental applications, while data-driven experimental models are not designed to infer global dynamic behaviors or frequency responses. This leaves a substantial gap in the intermediate systems that are experimentally realizable yet structurally simple, or derived from FEA. Bridging this gap requires gradually extending methods proven on simple models to systems of increasing physical complexity. Such an approach would enable the development of measurement-validated, interpretable surrogate models capable of extrapolating realistic frequency responses from limited vibration data, with direct relevance to component-level analysis in aerospace applications.

By the combination of neural Ordinary Differential Equations (ODEs) and neural operators, this work will demonstrate a machine learning algorithm with architectural modifications to allow extension into more complex problems by reducing the frequency band of training data required to extrapolate a frequency response curve for vibrating systems. Additionally, generalizations will be performed to make a workflow that allows arbitrary forcing functions to be applied to the system of inquiry. As a foundational proof of concept, this workflow will be tested on the damped linear oscillator, with the architecture designed to be compatible with future complexities of nonlinear, multiple degree-of-freedom, and experimental systems. By bridging the gap in machine learning algorithms, a more unified structure for analysis can be constructed. This is critically important to the improvement of design for components in mechanical vibration, as advanced modeling techniques reduce the design and experimentation iterations required. The ability to create a generalizable surrogate model also allows more realistic simulations to be performed, which avoids the difficulties of making experiments that capture all possible vibration conditions that could be encountered in the operation of an engineered system. In this work, a novel fusion of machine learning methods will be proposed with the following goals. First, increase the inference ratio of a machine learning algorithm on dynamical systems. Second, develop structures to generalize the solution of vibration problems accurately via machine learning. Third, analyze the results and properties produced by this new methodology, and identify the advantages and limitations on linear systems as a foundational proof-of-concept.

Section 2 will describe briefly the history and mechanisms of physics-based machine learning algorithms currently in the literature. Section 3 will introduce the theory and workflow for this study's algorithm, and Section 4 will analyze the algorithm's performance on a linear system and derive the relative coordinate system needed to execute base excitation. Stability analysis will be performed in Sections 5 and 6 to show how the algorithm behaves under the variance of the spectral range of training data, when forecasting higher frequency systems, and will characterize linear stability as training progresses between epochs. Finally, the future potential of this algorithm will be discussed, along with the larger problems that motivate its development.

2. **Related work in machine learning**

2.1. **Physics-Informed Neural Networks (PINNs)**

The history of physics-informed machine learning is a long one. The original methodology was the Physics-Informed Neural Network (PINN), developed by Raissi et al. [28]. The methodology of PINNs is to combine data-driven methods with the residual of a known governing differential equation to develop a model guaranteed to obey physical laws. First, a differential equation is posed:

$$u_t + \mathcal{N}[u; \lambda] = 0 \tag{1}$$

where $u(t, x)$ is the solution to the nonlinear operator $\mathcal{N}$ parameterized by $\lambda$.

PINNs can be used in two ways: a forward problem and an inverse problem. In the forward problem, a network is trained on the pairing of a dataset and the residual of a known differential equation. Through this combined training, the PINN becomes a model that respects the known governing equation and also the data provided. This is useful when the particular governing equation of a system is known and can simply be added to the loss function as a regularizing term. The second method of PINN usage is perhaps of greater interest to the experimentalist. Inverse PINN problems involve the discovery of the parameters of a governing differential equation by training on the dataset. In [28], the authors used this methodology to find the viscosity coefficient $v$ in Burger's equation. This is useful when the experimental dataset is known to obey a certain general form of Partial Differential Equation (PDE) or ODE, but its coefficients are unknown. It is important to note, as it was in [28], that without knowledge of the form of governing equation a system has, PINN application will be problematic. To use a PINN on a dataset is to assume all important parameters (effective mass, linear and nonlinear forms of damping, stiffness) are properly accounted for. PINNs can also fail to predict some instabilities and effects that appear in practice when the dataset is limited [29]. In practice, the PINN is well-suited to experimental problems with well-characterized physics, and can be useful to make a reliable, physically accurate model of a system.

### 2.2. Neural Ordinary Differential Equations (Neural ODEs)

Later, the Neural ODE was developed [30] as a model of a system's state space equations. In the original notation of the paper:

$$\frac{d\vec{h}(t)}{dt} = f_\theta(\vec{h}(t), t) \quad (2)$$

where the hidden layer $\vec{h}$ can be modeled at any layer depth $t$ by integration of the vector equation.

The benefit of Neural ODEs is that few assumptions are placed on the form of the system that it can identify. The process of reducing a system to state-space equations has an established validity, and is often a more convenient form to induce forecasts, making Neural ODEs a more streamlined type of physics-informed network. The architecture is compatible with a wide array of numerical methods [25], and in fact, the interaction of the neural network and numerical integration is central to operation. With modification, this system can be used to good effect on stiff systems. This is especially useful where the physics of differing timescales are present, like in combustion dynamics [27, 31]. The primary limitation of Neural ODEs is the simplicity of the neural architecture. Without modification, it does not readily lend itself to solutions in vibration, where nonlinearity, aperiodicity, and other physical effects complicate the map between the input and output sets.

### 2.3. Lagrangian Neural Networks (LNNs) and Hamiltonian Neural Networks (HNNs)

An important development in efforts to utilize machine learning on dynamics comes from the application of first-principle energy methods like the Euler-Lagrange principle and Hamiltonian principle. The foundational work on this idea began with the development of the Hamiltonian Neural Network [32], in which the position $p$ and momentum $q$ were employed as inputs to a feedforward network, which output the derivatives of these quantities, $\left(\frac{\partial p}{\partial t}, \frac{\partial q}{\partial t}\right)$. Next, another network was used to combine this data into the scalar Hamiltonian $\mathcal{H}$. For non-dissipative systems, the Hamiltonian represents total mechanical energy and does not change in time. Because of this, it allowed the network encoding to represent a system's motion in a way that implicitly obeyed physical principles. In this way, the task of modeling an undamped spring-mass system or pendulum system became more tenable via machine learning.

Of course, a significant limitation of a Hamiltonian representation is the pervasive presence of damping and dissipation in physical systems, which is the main source of motivation for the Lagrangian Neural Network. Based on the principle of least action, it is known that the Lagrangian of a system $\mathcal{L} = T - V$ at all times obeys the equality [33]:

$$\frac{d}{dt}\left(\frac{\delta \mathcal{L}}{\delta \dot{q}}\right) - \frac{\delta \mathcal{L}}{\delta q} = Q_{NC} \tag{3}$$

By taking the velocity and displacement of a system as inputs to a feedforward network, the Lagrangian can be modeled. In training, the need to quantify the kinetic, potential, and dissipation energies is avoided by application of [33]:

$$\hat{\ddot{q}} = \left(\frac{\partial^2 \mathcal{L}}{\partial \dot{q}^2}\right)^{-1} \left(\frac{\partial \mathcal{L}}{\partial q} - \dot{q}\frac{\partial^2 \mathcal{L}}{\partial q \partial \dot{q}}\right) \tag{4}$$

This methodology allows implicit physical obedience with respect to dissipation effects and has seen broad use in dynamics modeling. It has been used in the literature to model Euler-Bernoulli beam dynamics [34], computational fluid dynamics [35], and expanded to consider system constraints [36]. While energy-based networks have a high degree of fidelity on dynamic problems, the disadvantages mainly appear in the practicality of application. First is the computational expense of computing the Hessian matrix in equation (4), which is a problem it has in common with the application of higher-order convergence schemes for back-propagation [37]. Another drawback is that in practice, LNNs can struggle to accurately identify nonconservative forces [38]. In addition, LNNs on dynamics problems may have a large training data size requirement to reach convergence, as demonstrated in [34].

**2.4. Deep Operator Networks (DeepONet)**

One of the latest developments in physics-based machine learning was the operator network. Operator networks first gained abilities in PDE solution, with the Deep Operator Network (DeepONet) design later allowing general nonlinear mapping between spaces [44]. Operator networks are centered around the use of an encoding kernel. These kernels specialize in layers of a network to represent different mathematical structures based on the operations performed on them. For example, Fourier Neural Operators (FNOs) [40] utilize a Fourier kernel by extracting the frequency information of a signal and using it to provide a nonlinear map between input and output data. DeepONet generalizes this by use of a bifurcated network structure, consisting of a branch network, which encodes the dependent variable of a system $u$ by $G(u)$, and an independent variable encoding trunk network to represent the variable $x$. These representations are merged to produce $G(u)(x)$, which is mathematically represented [39]:

$$G(u)(x) = \sum_{l=1}^{p} b_l t_l \tag{5}$$

where $l \in [1, p]$ is the layer depth of the network, $b_l$ denotes the state-encoding branch output of the $l^{th}$ layer, and $t_l$ is the trunk coordinate encoding of the $l^{th}$ layer.

Together, the branch and trunk networks allow the learning of an operator function $G(u)(x)$, which is a map between the input and the output spaces. This method allows better generalization for the same dataset size than Neural ODEs, making it a more data-efficient method than a naïve network implementation on dynamical data. Most expansions of the use of DeepONet have been on PDEs, and in particular, the ability to perform one-shot learning on systems like Darcy flow fields and Navier-Stokes problems [39, 40]. The original formulation of the DeepONet [39] proposes an input-output scheme that involves parallel computation of the solution space; that is, the solution is not executed by a Markovian process, even if the underlying system is Markovian in nature. This presents a problem in vibration analysis of ODEs, but not in PDEs, because the spatial domain of numerical PDE problems is inherently restricted, and often the timescale of interest is not a long one.

Vibration problems require high accuracy over long timescales to reach the steady-state solution of interest. The timescale at which this solution is achieved varies from problem to problem, and can only be known from FEA or experimentation for uncharacterized systems. The DeepONet, in its common formulation, however, requires set sensors in the independent coordinate space and returns solutions only valid for those locations. Generalization to other forms of the same system is the focus of the original formulation, but extrapolation to any timescale is problematic, especially when the given dataset does not extend to the desired forecast length. It is for this reason that the vanilla DeepONet is not ideal for vibration problems.

### 3. Delta Implicit Neural Operator (DINO) methodology

The methodology introduced here is a fusion of the Neural ODE and the neural operator. First, rather than using direct state flow, the DINO is posed as a delta operator like the Neural ODE. The objective is to identify the gradient space of an ODE given the state and driving excitation of the system at a given time with the use of an operator function. In DINO 1.0, this is done naively, passing state, time, and driving acceleration into the network and forecasting using the explicit forward Euler method.

$$\tilde{G}(\vec{x}, t, u) = B(\vec{x}, u) \cdot T(t) \tag{6}$$

$$\vec{x}^{k+1} = \vec{x}^k + \Delta t \cdot \tilde{G}(\vec{x}^k, t^k, u(t^k)) \qquad (7)$$

where $\tilde{G}(\vec{x}, t, u)$ denotes the combined operator composed of branch network $B(\vec{x}, u)$ and trunk network $T(t)$, with arguments of state $\vec{x}$, time $t$, and forcing function $u(t)$.

From here on, all network quantities and predictions will be denoted with a tilde to represent a localized gradient encoding. In subsequent versions, this method is modified to avoid the network directly encoding the driving function, focusing instead on natural system physics. This is done by bifurcating the network into an amplitude and phase branch set, and isolating the free response

$$\tilde{G}(\vec{x}) = A(\vec{x}) \cdot \Phi(\vec{x}) \qquad (8)$$

$$\vec{x}^{k+1} = \vec{x}^k + \Delta t \cdot \left[ \tilde{G}(\vec{x}^k) + \begin{Bmatrix} 0 \\ u(t) \end{Bmatrix} \right] \qquad (9)$$

where $A(\vec{x}), \Phi(\vec{x})$ are amplitude and phase branches used to create separated scalar and vector latent spaces in the network.

Note that this assumes the form of the ODE to be linearly dependent on the driving function (i.e., the ODE does not contain a term that involves products of the driving function and dependent variable). For most well-posed vibration problems, this assumption is valid. This is an advantageous alteration, as it allows far more generality in the driving function, and allows the DINO to act more as a surrogate for the global behavior of a system, rather than an extrapolation of a single condition. The most common notable exception where this formulation does not work is in parametrically driven oscillators, like the Mathieu-Duffing equation [41]. In these systems, work is applied to the system such that the nonhomogeneity is coupled to the dependent variable. In practice, this means that the system's excitation is not simply superimposed upon the state space equations, which reduces the generality of the DINO algorithm

**3.1. Model system description: Universal oscillator equation (LS-1)**

Figure 1 shows the envisioned workflow to be applied to test systems. First, a limited amount of harmonic testing data is obtained, and then characterized with machine learning. Next, the dynamical system is tested for stability and extrapolated in the frequency domain to generate a frequency response curve. This process should begin with a simple proof of concept that is widely applicable. To this end, in this work, the main focus will be on a linear spring-mass-damped subjected to harmonic forced or base excitations.

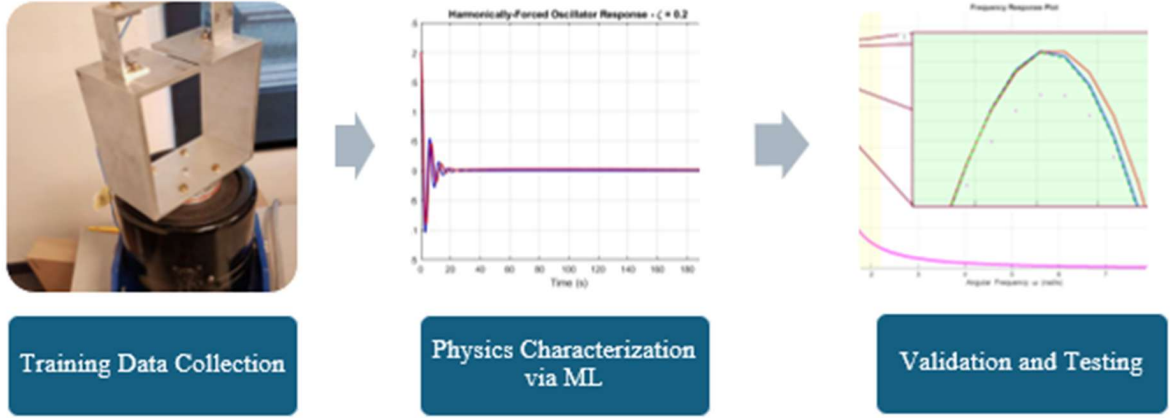

Figure 1. Machine Learning (ML) workflow for vibration analysis.

The harmonically forced linear oscillator is considered the most basic vibrating system, due to its linear damping and restoring force. As such, it is a good preliminary choice for testing the DINO on vibration forecasting. Its equation is:

$$\ddot{x} + 2\xi\omega_n\dot{x} + \omega_n^2 x = A\cos(\omega t) \tag{10}$$

where $\omega_n$ represents the natural frequency of the system in $\frac{rad}{s}$, determined by its mass and stiffness, $\xi$ is the damping ratio, $A$ is the driving acceleration excitation in $\frac{m}{s^2}$, and $\omega$ is the driving frequency in $\frac{rad}{s}$. While values of mass, stiffness, and damping can be selected for physical realism, it is convenient for initial validation to perform nondimensionalization as in [42] to arrive at the universal oscillator equation.

First, make the definitions:

$$r = \frac{\omega}{\omega_n}, \tau = \omega_n t, q = \frac{x}{L}$$

Next, redefine (10) using the nondimensionalizing terms:

$$L\omega_n^2\ddot{q} + 2\xi L\omega_n^2\dot{q} + L\omega_n^2 q = A\cos(r\tau) \tag{11}$$

With $L = \frac{A}{\omega_n^2}$, the term $L\omega_n^2$ can be eliminated:

$$\ddot{q} + 2\xi\dot{q} + q = \cos(r\tau) \tag{12}$$

Two forcing systems will be studied here: forced excitation and base excitation. In most experimental designs for vibration analysis, base excitation is more commonly used than forced excitation. Base excitation is the application of a known displacement to a shaker or other support fixture at the base of the system of interest. To be fully relevant to experimental datasets and have practical application, the DINO algorithm must be made compatible with this type of excitation. DINO results for this system will be shown in Section 4.3.

Based on the schematic shown in Figure 2, the dimensional governing equations of motion of the base-excited spring-mass-damper can be written as:

$$m\ddot{x} + c\dot{x} + kx = c\dot{y} + ky \tag{13}$$

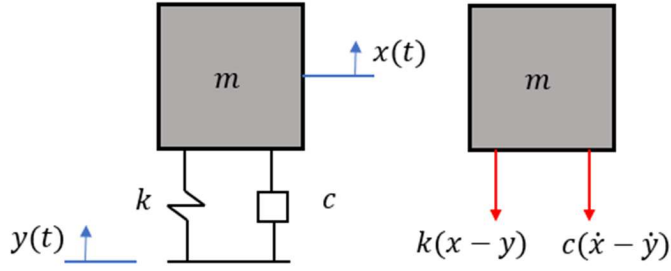

Figure 2. Configuration of the base excitation model.

Data generated by the universal oscillator equation (equation (12)) is denoted as the LS-1 training dataset. This data has $\xi = 0.2$, $r_p = 0.979$. The training curriculum is generated by choosing a bandwidth of driving frequencies within the desired Frequency Response Curve (FRC) to be forecast, and randomly selecting ten forced trajectories. These trajectories each have a randomized initial condition for position and velocity, which is chosen out of ten Banded Random Uniform (BRU) bands within a specified range from 0.001 to 1.0. This allows the network to learn physics as the system amplitude gradually increases. In DINO 1.0, the training algorithm can be described as:

$$\theta^* = \underset{\theta \in \Theta}{\mathrm{argmin}} \frac{1}{N} \sum_{k=1}^{N} \mathcal{L}\left(\tilde{G}(x^k, t^k, u(t^k)), f(x^k, t^k, u(t^k))\right) \tag{14}$$

Here, the outputs of the state transition function and operator function are directly compared using L1 loss, a neural network modification of Mean Absolute Error (MAE). In subsequent versions, decoupling the driving function from network input requires a modified training algorithm.

$$\theta^* = \underset{\theta \in \Theta}{\mathrm{argmin}} \frac{1}{N} \sum_{k=1}^{N} \mathcal{L}\left(\tilde{G}(\vec{x}^k), f(\vec{x}^k, t^k, u(t^k)) - u(t^k)\right) \tag{15}$$

Subtracting the driving excitation from the state transition, the operator function is trained on harmonically excited data with the forcing acceleration $u(t)$ subtracted. Unlike equation (14), this design allows a more robust encoding, and also allows arbitrary forcing types like rotating unbalance, base excitation, random excitation, and impact excitation to be performed on the same system purely by inference.

### 3.2. Numerical integration of neural network quantities

The function of the DINO system is to form a surrogate of the state transition function. As a result, it returns the gradient space of an ODE. For this reason, the use of a numerical method is necessary to determine the time response and frequency response curve of the system in terms of position. The interaction of the network and numerical method is essential to performance. In the testing performed, recursive prediction is used to interrogate a time history with a 4$^{th}$ order Runge-Kutta (RK4) scheme nested in an implicit trapezoidal scheme. By substitution of the state transition function with the neural operator function, the scheme can be expressed as [43]:

$$\begin{cases} y^*_{i+\frac{1}{2}} = y_i + \frac{h}{2}\tilde{G}(x_i, y_i) \\ y^{**}_{i+\frac{1}{2}} = y_i + \frac{h}{2}\tilde{G}\left(x_{i+\frac{1}{2}}, y^*_{i+\frac{1}{2}}\right) \\ y^*_{i+1} = y_i + h\tilde{G}\left(x_{i+\frac{1}{2}}, y^{**}_{i+\frac{1}{2}}\right) \\ y^{**}_{i+1} = y_i + \frac{h}{6}\left[\tilde{G}(x_i, y_i) + 2\tilde{G}\left(x_{i+\frac{1}{2}}, y^*_{i+\frac{1}{2}}\right) + 2\tilde{G}\left(x_{i+\frac{1}{2}}, y^{**}_{i+\frac{1}{2}}\right) + \tilde{G}(x_{i+1}, y^*_{i+1})\right] \end{cases} \quad (16)$$

For additional stability, RK4 is used as an initial guess for the implementation of the implicit trapezoidal scheme. This allows for greater stability than an explicit method, and allows a useful manipulation of neural network properties to accelerate convergence.

The trapezoidal stencil [43] is:

$$\vec{x}^{k+1} = \vec{x}^k + \frac{h}{2}\left[\tilde{G}(\vec{x}^k, \vec{u}^k) + \tilde{G}(\vec{x}^{k+1}, \vec{u}^{k+1})\right] \quad (17)$$

From which the residual is defined:

$$\vec{R} = \vec{x}^{k+1} - \vec{x}^k - \frac{h}{2}\left[\tilde{G}(\vec{x}^k, \vec{u}^k) + \tilde{G}(\vec{x}^{k+1}, \vec{u}^{k+1})\right] \quad (18)$$

Then, take the Jacobian of the operator function by automatic differentiation of the network:

$$\frac{\partial \vec{R}}{\partial \vec{x}} = \begin{bmatrix} \frac{\partial R_1}{\partial x_1} & \frac{\partial R_1}{\partial x_2} \\ \frac{\partial R_2}{\partial x_1} & \frac{\partial R_2}{\partial x_2} \end{bmatrix} \quad (19)$$

And use the Newton-Raphson update rule:

$$\vec{x}^{k+1} \leftarrow \vec{x}^{k+1} + \left(\frac{\partial \vec{R}}{\partial \vec{x}}\right)^{-1} R \quad (20)$$

The forecast is then iterated to the desired convergence. Using an implicit $O(h^2)$ scheme with a nested $O(h^4)$ explicit initial forecast provides a guaranteed numerical stability at a higher computational cost. It should be noted that because this scheme is performed on a neural network rather than a known set of state space equations, convergence of the final solution is not obligatorily guaranteed, but the numerical scheme itself is stable when the necessary conditions are satisfied.

The largest time step for which the DINO may be successfully trained is governed by the Shannon-Nyquist Sampling Theorem [31].

$$f_R = \frac{1}{\Delta t_{max}} = 2f_{target} \tag{21}$$

where $f_R$ is the Nyquist rate, $f_{target}$ is the maximum frequency in the solution bandwidth, and $\Delta t_{max}$ represents the largest allowable step size.

The minimum sampling rate to capture all spectral data in a region less than the target frequency is twice that target frequency. For the LS-1 system, $f_R = 3.18$ is used to capture the entire frequency response curve of interest from 0.1 to 10 dimensionless frequency units. This corresponds to a maximum training time step of $\Delta t_{max} = 0.314$. Any higher value will make the signal subject to aliasing and invalidate training. In practice, numerical methods have very poor performance at this sampling rate, so a time step of 0.01 is used in all nondimensional LS-1 tests, resulting in a sampling rate 62.83 times higher than the maximum target frequency.

## 4. Testing and analysis of the DINO algorithm

### 4.1. LS-1 forced excitation time history and frequency response: Versions 1.0 - 3.0

All time histories are queried from the network under a driving frequency $r = 3.77$, and initial conditions $q_0, \dot{q}_0 = 0.2, 0$, respectively, with a standard LS-1 set of system parameters ($\xi = 0.2$). This driving is in the post-resonant region of the system, allowing the performance to be clarified on higher frequencies. Frequency responses are computed by parallel inference of 500 trajectories, using the Hilbert amplitude of each trajectory to identify the steady-state amplitude. These are forced using frequencies in the band of $0.1 - 10.0$, with a driving amplitude of $1.0$. Because of the character of the simulation, this is neither an upsweep nor a down-sweep, but a discrete testing of every separate scenario encompassed in the frequency response curve. In the development of the DINO system, nine configurations were designed to realize stability, accuracy, or accelerated training performance. As shown in Figure 3, three versions are to be shown here to showcase the flow of structural changes and their effect on the properties of the forecast.

DINO 1.0, shown in Figure 3(a), consisted of an irregular 1:4 activation layer ratio in the branch network to enforce better encoding. The trunk network had a 1:2 activation layer ratio. The normal distribution of initial conditions undertrained higher amplitudes, causing limited capture of initial conditions. This network takes a dual tensor input of [batch, 3] (state and driving) with [batch, 1] (normalized time) during training, and [1, 3] during inference, receiving normalized position, velocity, and driving acceleration. Additionally, DINO 1.0 was modified to use the staged fourth-order Runge

Kutta (RK4) integrator paired with an implicit trapezoidal scheme for forecasting. This version also employs biasing of the branch network. It was theorized that this would reduce amplitude error and discourage instability in the output.

In the design of DINO 2.0, presented in Figure 3(b), several possible causes of inaccuracy were investigated. During experimental modifications, the trunk network's time encoding was identified as the primary cause of the errors in the frequency response curve. DINO 1.0 as a whole was nonautonomous because of the presence of normalized driving in the system input. But due to the nature of the nonautonomous system, the trunk network encoded normalized time to prevent saturation of the activation functions. These inputs both inhibit the response accuracy (as time cannot be satisfactorily normalized in inference without making it cyclical) and limit the selection of a driving function to the one used to train the network. For these reasons, the forcing function was removed from the input and added to the output of the network, and the time encoding was removed, as shown in Figure 3(b). In addition to the prospect of improving accuracy, these modifications also allow the time resolution and forcing function to be selected upon inference as any arbitrary values, regardless of the training data used, which aids the generality of the system. In DINO 3.0, the state input is altered to a $[\beta, 2\nu]$ tensor, where $\beta$ denotes the batch size and $\nu$ is the number of degrees of freedom. Additionally, the state encoding is bifurcated into an amplitude and phase branch, as indicated in Figure 3(c) instead of a state-time encoding. Forcing is added to the network output, so the network trains on the autonomous component of the ODE.

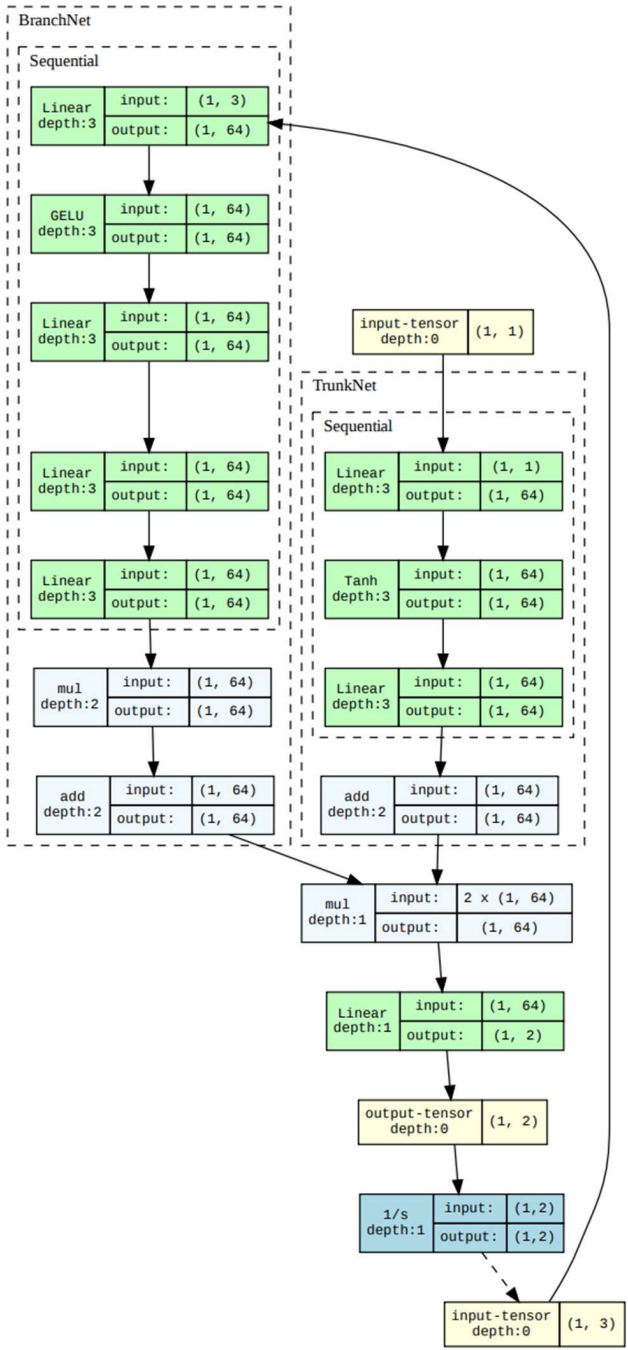
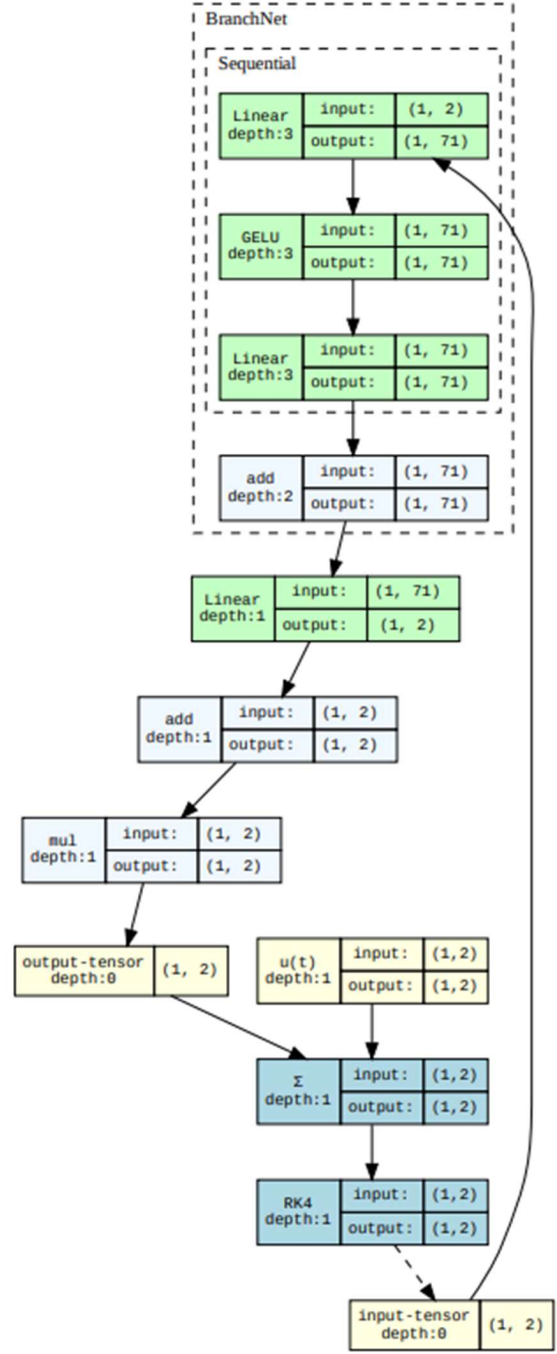

(a) (b)

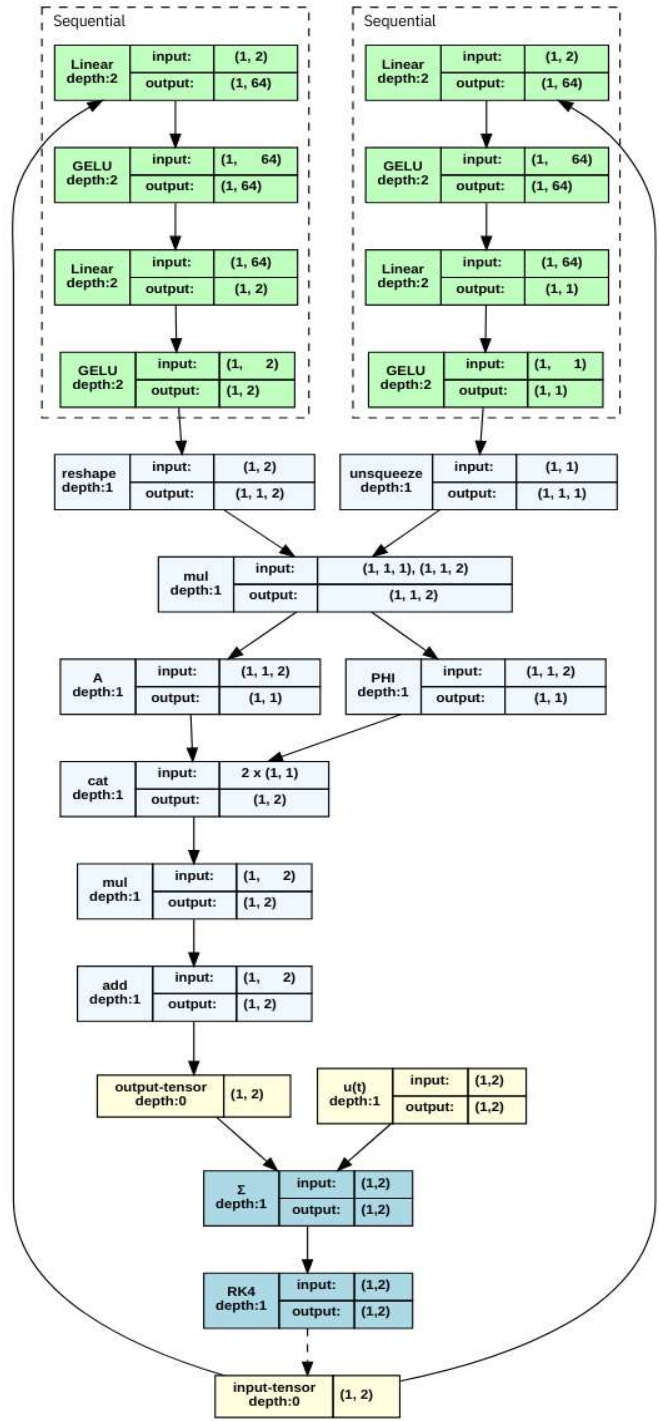

(c)

Figure 3. Comparison of (a) DINO 1.0, (b) DINO 2.0, and (c) DINO 3.0 network architectures.

As indicated in the plotted time histories in Figures 4(a-c), the DINO 1.0 time response shows a good qualitative match. Mean shift, amplitude error, and phase error dominate the error of the response, the largest of them being amplitude at 13.3% on average between the BRUs tested, as shown in Table

1. The phase error represents an overprediction of 9.57% on average, with a consistent offset to the true trajectory. This suggests that the offset is stable in the neural network representation. The error is generally uniform, showing the same inaccuracies in the transient and steady-state regimes. Further, tests indicate network stability on long time histories of 100, beyond the uniform training horizon (shown in yellow). The 0.1-0.8 BRU shows the highest amplitude error, and an additional mean shift greater than that of resonantly-trained and post-resonantly trained networks. The overall accuracy of DINO 1.0 in the time response is 81.91%.

    The modifications of DINO 2.0 are shown in Figures 4(d-f) and Table 1 to increase the overall time response accuracy to 99.60%, which indicates that removing time encoding has the expected effect on convergence. DINO 2.0 shows a significant 5% reduction in phase error, going from an overprediction to a slight underprediction. In comparison, DINO 3.0 shows a near-complete convergence in Figures 4(g-i), with an overall accuracy of 99.99% in the time response. The greatest error in the DINO 3.0 response is the amplitude component, with all other metrics being below 1.00%. An interesting break with BRU curricula in previous versions is that the DINO 3.0 error metrics are far more comparable between bandwidth selections, indicating that the amplitude-phase encoding can more readily converge on off-resonance time histories. It appears from the information in Table 1 that pre-resonant training data leads to the most accurate time history reconstruction, though it will be shown later, in Section 5.1 that BRU band location selection must be done to realize the best global frequency response curve, not simply to maximize convergence of a single frequency's time history. Further, the errors of Table 1 will be contextualized in Section 5.3 to show how the root task of the DINO algorithm is to identify the eigenvalues of a vibrating system on a real-imaginary plane. Studying the training convergence in this way can be instructive to explain why the error components appear the way they do in these results.

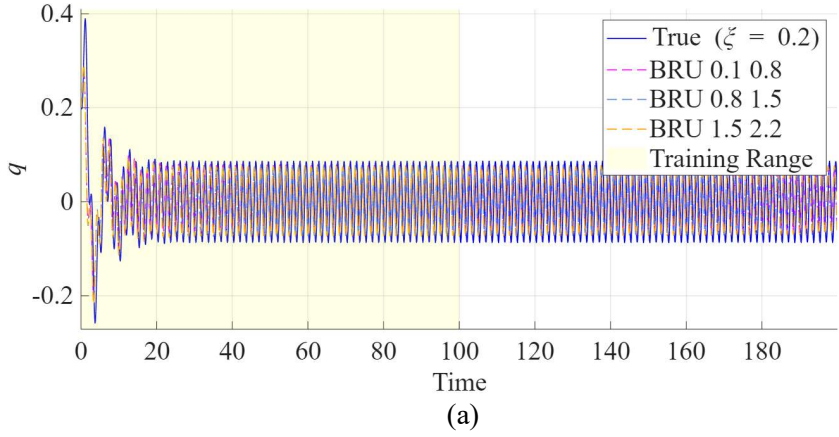
(a)
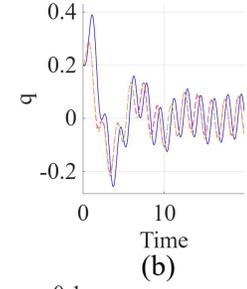
(b)
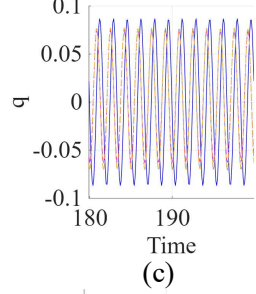
(c)
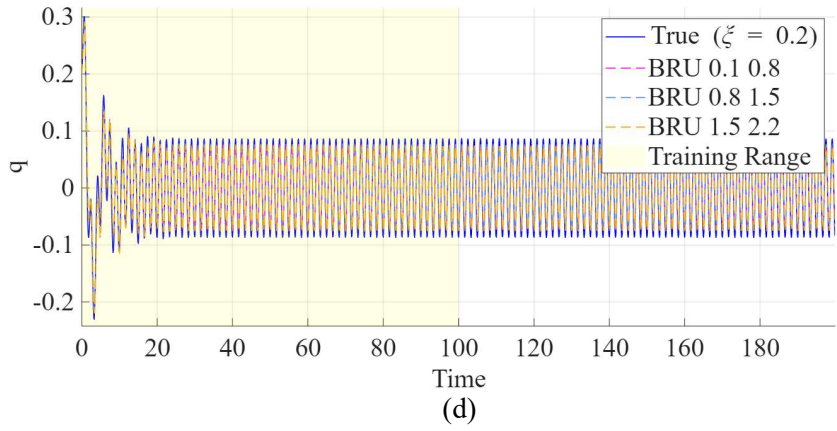
(d)
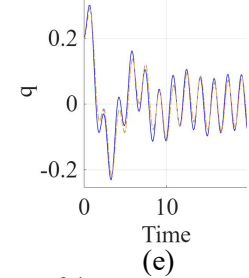
(e)
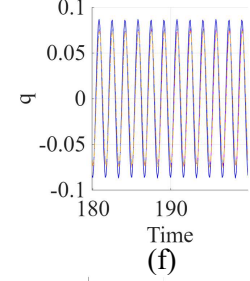
(f)
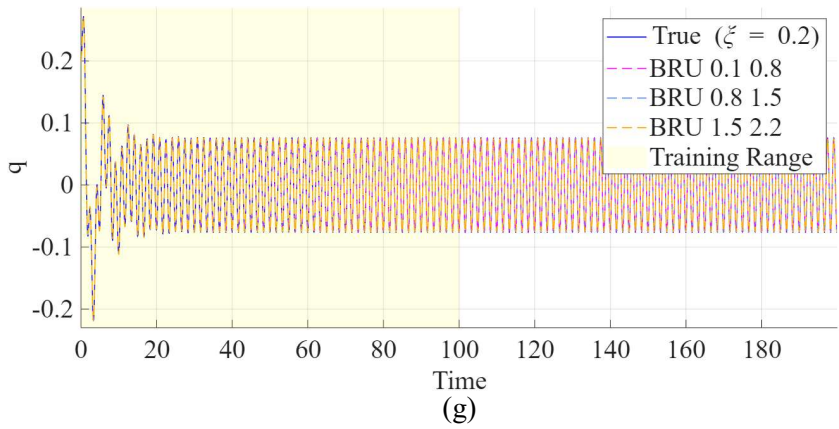
(g)
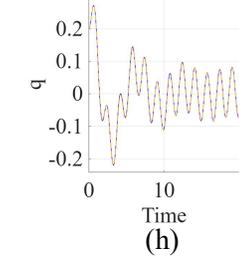
(h)
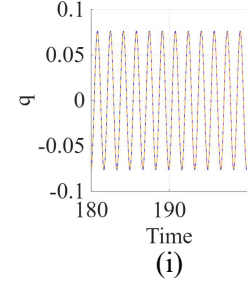
(i)

Figure 4. (a) DINO 1.0 inference results in the time domain, (b) transient, and (c) steady state on the universal oscillator equation with training regions (BRUs) colored magenta, cyan, and yellow with increasing frequency band range. (d-f) show the results of DINO 2.0 inference, and (g-i) show DINO 3.0 results. Time history generated at a driving condition of $r$=3.77.

Table 1. Time response accuracies of DINO 1.0, 2.0, and 3.0 compared among three BRU bands. Bottom rows denote percentage accuracy, while middle rows represent percentage error.

| DINO 1.0 time response accuracy | | | | | |
|---|---|---|---|---|---|
| BRU | MSE | Mean error (%) | Amp. error (%) | Phase error (%) | Freq. error (%) |
| 0.1-0.8 | 5.47E-03 | 10.38 | 13.92 | 9.32 | 1.02 |
| 0.8-1.5 | 5.47E-03 | 22.99 | 13.04 | 9.68 | 1.04 |
| 1.5-2.2 | 5.52E-03 | 20.90 | 13.04 | 9.71 | 0.97 |
| Accuracy (%) | | 81.91 | 86.67 | 90.43 | 98.99 |

| DINO 2.0 time response accuracy | | | | | |
|---|---|---|---|---|---|
| BRU | MSE | Mean error (%) | Amp. error (%) | Phase error (%) | Freq. error (%) |
| 0.1-0.8 | 2.35E-04 | 0.40 | 13.24 | 4.10 | 0.95 |
| 0.8-1.5 | 2.38E-04 | 0.40 | 13.12 | 4.12 | 0.94 |
| 1.5-2.2 | 2.32E-04 | 0.40 | 13.16 | 4.09 | 0.91 |
| Accuracy (%) | | 99.60 | 86.83 | 95.90 | 99.07 |

| DINO 3.0 time response accuracy | | | | | |
|---|---|---|---|---|---|
| BRU | MSE | Mean error (%) | Amp. error (%) | Phase error (%) | Freq. error (%) |
| 0.1-0.8 | 1.80E-06 | 0.01 | 1.16 | 0.08 | 0.21 |
| 0.8-1.5 | 1.91E-06 | 0.01 | 1.19 | 0.08 | 0.21 |
| 1.5-2.2 | 1.96E-06 | 0.01 | 1.21 | 0.08 | 0.22 |
| Accuracy (%) | | 99.99 | 98.81 | 99.92 | 99.78 |

As shown in Figure 5(a), the loss curve on a 100-epoch training run is monotonic for BRUs 2 and 3, but not for BRU 1, which may explain the poorer time response of that BRU in DINO 1.0. In contrast to the DINO 1.0 test, all loss curves in Figures 5(c) and 5(e) for DINO 2.0 and DINO 3.0 are monotonically decreasing, indicating the improved consistency of results across all BRUs tested. The training was altered in DINO 2.0 and DINO 3.0 to enable plateau-breaking through PyTorch's ReduceLROnPlateau function. A general trend seen in this data is that higher-frequency BRUs result in plateaus at lower epoch counts, indicating a faster convergence. Looking at the phase space trajectories in Figures 5(b, d, f), it can be noted that the position inaccuracy between the true solution and DINO solutions is greater than that of the velocity inaccuracy. This can be explained by the way the model is applied to induce a forecast by numerical integration of the neural operator. When integrated via staged RK4 and implicit trapezoidal methods, the truncation error of these schemes causes the accuracy of the prediction to degrade. Velocity and acceleration are directly trained for in the output vector during back-propagation, but the position is only correlated to the values of velocity and acceleration at a given state.

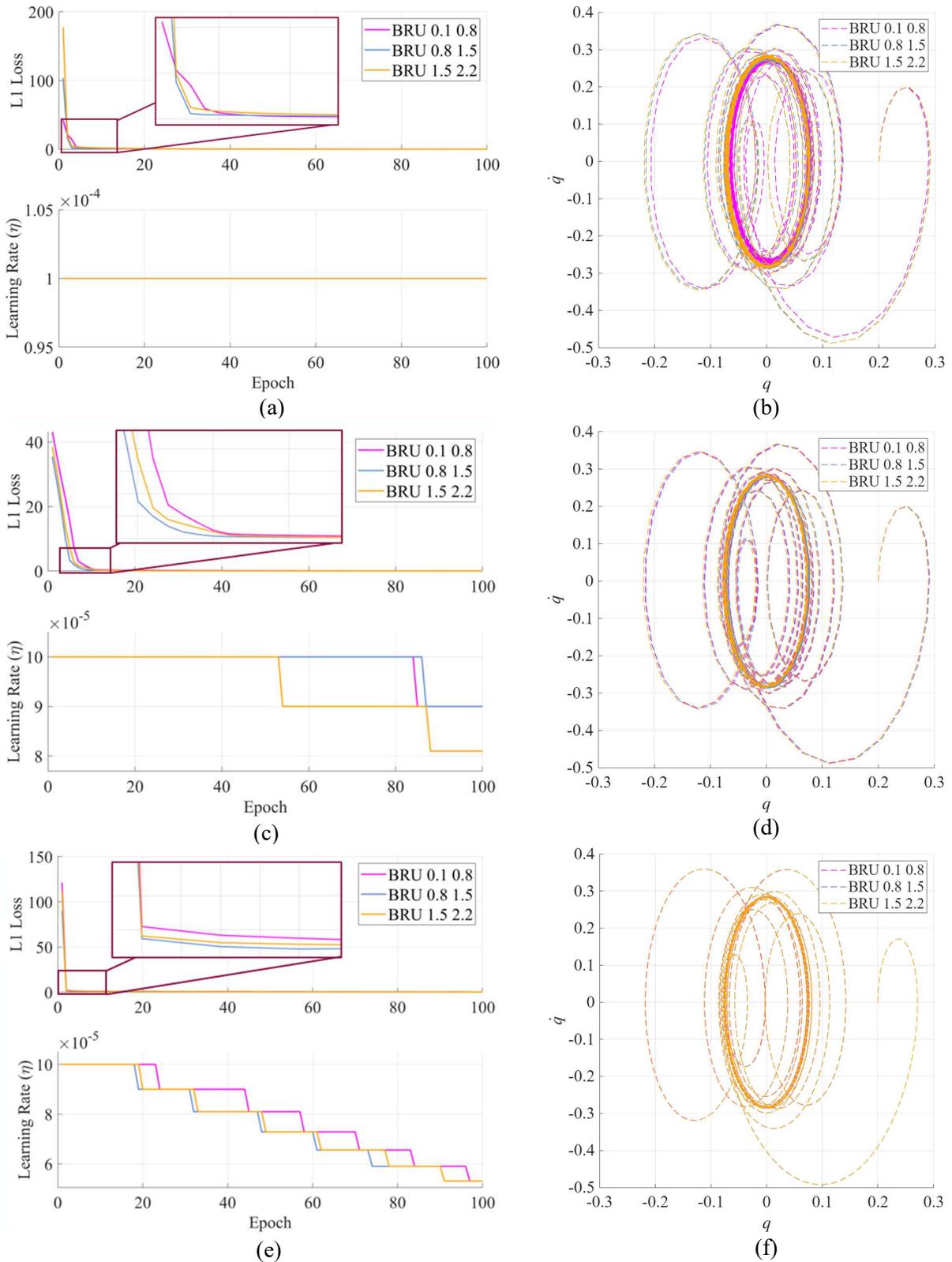

Figure 5. (a, c, e) L1 loss and learning rate over 100 epochs of training for DINO 1.0, 2.0, and 3.0 and (b, d, f) phase space trajectory of BRUs 1, 2, and 3.

The next step of the workflow is to use parallelized computation to realize the DINO frequency response curve estimates for the regime of 0.1-10. As with the time histories, three BRU curricula will be compared to draw conclusions about how the bandwidth selection affects performance. Despite the passable accuracy of the DINO 1.0 time responses denoted in Figures 4(a-c) in all BRU regions, the frequency response curve of the DINO 1.0 shows a lower accuracy than expected, as indicated in Figure 6(a). The frequency response curves globally underpredict amplitude at all frequencies and BRU bands, and suggest a low-frequency resonance at approximately 0.55. The start and endpoints of the true and predicted responses are generally matched, suggesting a limited physicality of the solutions during frequency extrapolation. The most significant inaccuracy of the results is the resonance shift, suggesting that phase error must be targeted to realize better performance, as shown in Figure 6(a) and Table 2.

The DINO 2.0 frequency response curve shows a far better alignment and physicality across all BRUs, and a near-perfect response in off-resonance regions. The sole sources of error in this response are in the amplitude of the peak, which is less than 5% across all BRUs, as indicated in Figure 6(b) and Table 2. Moreover, the identified resonant frequency of the system is fully accurate to float32 (7-decimal place) machine precision, the highest accuracy achievable by the network specifications. The lowest accuracy is found in the pre-resonant BRU 1, followed by BRU 2. As expected, the BRU which contains resonance is the most accurate globally. This corrects any faulty assumptions insinuated by the time responses of Figure 4, it does hold true that resonance contains the richest dynamical data, even if off-resonance training may locally enhance time histories at certain frequencies.

DINO 3.0 results in Figure 6(c) and Table 2 indicate that the main improvement of the amplitude-phase encoding in the frequency domain is the reduction of peak amplitude error. As with DINO 2.0, DINO 3.0 has full machine precision identification of the frequency where the peak occurs, and shows an overall accuracy of 99.87% between all BRUs tested. It is promising to note that the DINO 3.0 response has 100% accuracy on the location of resonance, if not its amplitude, indicating a high-degree of clarity in the identification of the frequency and damping characteristics of the universal oscillator equation.

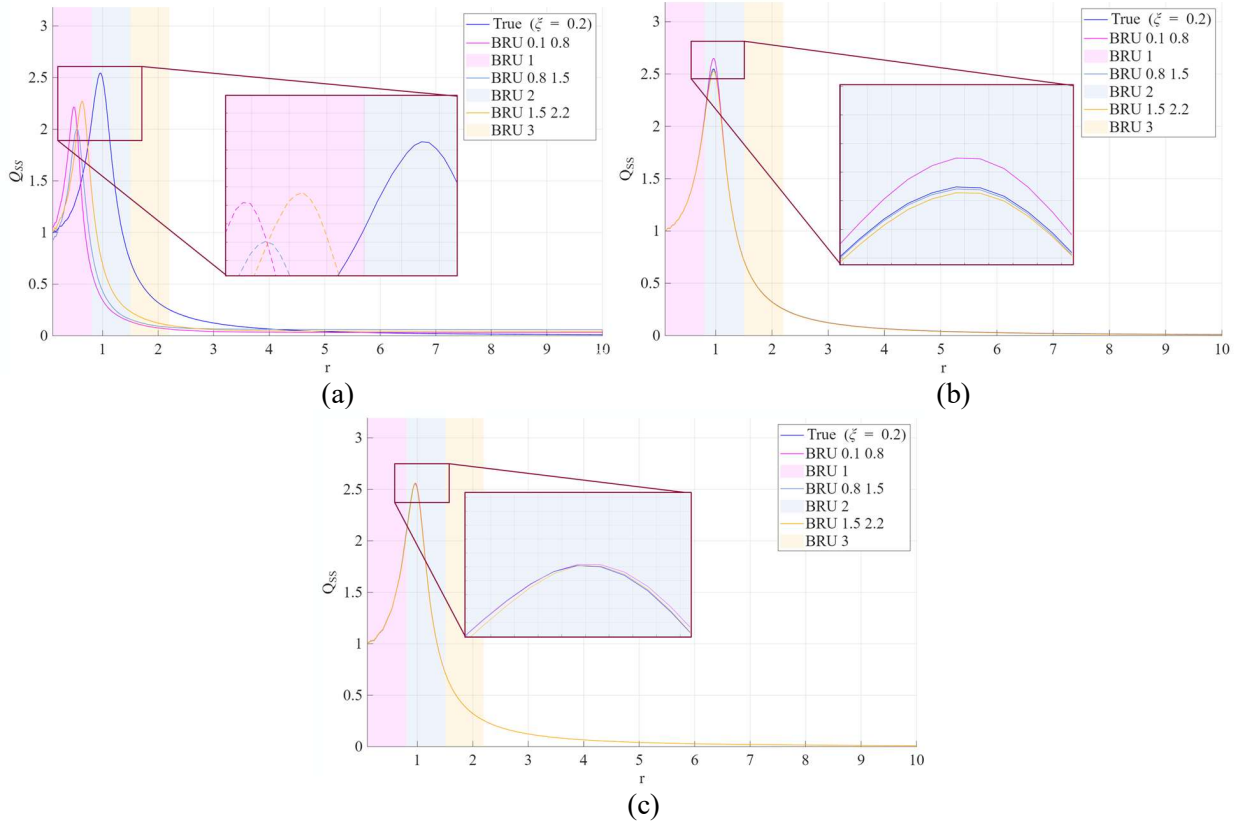

Figure 6. Output displacement versus the excitation frequency for (a) DINO 1.0, (b) DINO 2.0, and (c) DINO 3.0.

Table 2. Frequency response accuracy DINO 1.0, 2.0, and 3.0.

| DINO 1.0 frequency response accuracy | | | | |
|---|---|---|---|---|
| BRU | Shape error (%) | Peak error (%) | Resonance error (%) | $\hat{r}_p$ |
| 0.1-0.8 | 17.23 | 12.97 | 49.96 | 0.48 |
| 0.8-1.5 | 1.55 | 21.22 | 43.71 | 0.54 |
| 1.5-2.2 | 6.57 | 10.72 | 33.31 | 0.64 |
| Accuracy (%) | 91.55 | 85.03 | 57.67 | 0.55 |

| DINO 2.0 frequency response accuracy | | | | |
|---|---|---|---|---|
| BRU | Shape error (%) | Peak error (%) | Resonance error (%) | $\hat{r}_p$ |
| 0.1-0.8 | 7.82 | 3.95 | 0.00 | 0.95 |
| 0.8-1.5 | 6.01 | 0.28 | 0.00 | 0.95 |
| 1.5-2.2 | 6.18 | 0.81 | 0.00 | 0.95 |
| Accuracy (%) | 93.33 | 98.32 | 100.00 | 0.95 |

| DINO 3.0 frequency response accuracy | | | | |
|---|---|---|---|---|
| BRU | Shape error (%) | Peak error (%) | Resonance error (%) | $\hat{r}_p$ |
| 0.1-0.8 | 0.12 | 0.16 | 0.00 | 0.95 |
| 0.8-1.5 | 0.02 | 0.06 | 0.00 | 0.95 |
| 1.5-2.2 | 0.26 | 0.02 | 0.00 | 0.95 |
| Accuracy (%) | 99.87 | 99.92 | 100.00 | 0.95 |

In summary, the progression of DINO 1.0, DINO 2.0, and DINO 3.0 represents the adaptation of the network architecture to capture the state-space dynamics of an unknown system by providing a limited bandwidth of harmonic vibration testing. A performance description of the DINO would be incomplete without a discussion of its time and computational requirements. All DINO calculations were run using GPU operations for all training and network inference tasks using an Nvidia A2000 graphics card. CPU operations were run on an Intel 13$^{th}$ Generation i9, with a clock rate of 3.00 GHz. Due to the obligatorily serial nature of the forecast, inference tasks can only be accelerated by parallel inference of multiple trajectories. Frequency response curves of the dimensionless displacement are calculated in this way using 500 parallel trajectory inferences on the GPU. The time required for training was reduced in total by the use of driving force cancellation and elimination of the time encoding, as shown in Figure 7. Likewise, even with the DINO 3.0 model having a larger parameter count than the other two, its execution time still realizes a reduction in training time. The true drawback of using a larger model is the additional time required to infer a forecast. The primary time increase in the frequency response calculation is incurred by storing the large dataset of predicted trajectories. A summary of the time requirement percentage for each category is shown in Table 3. With respect to the total time needed for each model to perform the simulations, it is clear that the training time is reduced for both DINO 2.0 and DINO 3.0.

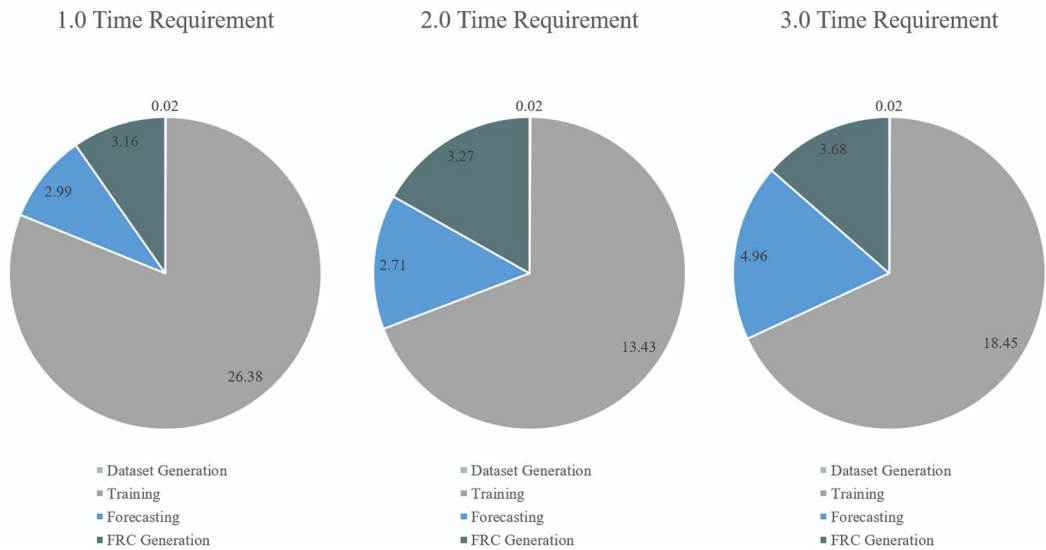

Figure 7. Time benchmarking for DINO 1.0-3.0 (quantities denoted in seconds of execution).

Table 3. Time requirement summary for all DINO versions.

|  | DINO 1.0 | DINO 2.0 | DINO 3.0 |
|---|---|---|---|
| Dataset generation (%) | 0.06% | 0.10% | 0.07% |
| Training (%) | 81.04% | 68.79% | 68.06% |
| Forecasting (%) | 9.18% | 13.88% | 18.30% |
| FRC calculation (%) | 9.71% | 16.75% | 13.57% |

**4.2. Dimensional LS-1 testing**

The previous section validates performance on LS-1 when the terms are nondimensional, allowing the natural frequency to be unity from a simulation point of view. While a good proof of concept, the next step is to use the dimensional system with a physically realistic stiffness. Two cases will be shown for dimensional solutions selected for resonant amplitudes of 2 cm (LS-1A) and 4 cm (LS-1B), possessing natural frequencies of 11.3 $\frac{rad}{s}$ and 7.99 $\frac{rad}{s}$, respectively, with a uniform damping ratio of 0.2. These solutions will be calculated with DINO 3.0 to demonstrate the effect of increased stiffness on the performance of the model. In this study, the system's performance must retain the accuracy of the universal oscillator equation tests on high-frequency dimensional systems when the sampling frequency ratio $\frac{f_s}{f_n}$ is held constant between these two tests, which is equal to 62.83. Considering this, to hold this ratio constant, the LS-1A and LS-1B tests must have step sizes of 0.0088 and 0.0125 seconds during training, respectively. To generate a realistic simulation, these dimensional systems are selected to have the properties described in Table 4.

Table 4. System description for LS-1A and LS-1B dimensional test cases.

|  | Configuration A | | Configuration B | |
|---|---|---|---|---|
| Damping ratio | 0.2 | | 0.2 | |
| Natural frequency (rad/s) | 11.30 | | 7.99 | |
| Peak amplitude (cm) | 2 | | 4 | |
| BRU 1 (rad/s) | 7.00 | 7.70 | 4.50 | 5.20 |
| BRU 2 (rad/s) | 10.50 | 11.20 | 7.50 | 8.20 |
| BRU 3 (rad/s) | 12.00 | 12.70 | 9.00 | 9.70 |
| Time step (sec) | 0.0088 | | 0.0125 | |
| Test driving frequency (rad/s) | 10.53 | | 7.45 | |

The time response, when the step size is matched with the previous test to retain a nominal frequency ratio, retains a good qualitative match in both transient and steady-state regimes, as presented in Figure 8. The phase lag found in this solution is analogous to those found in nondimensional LS-1 tests, but it is of a lower relative magnitude than the true solution. Per the data illustrated in Table 5, the average accuracy of the DINO on lower stiffness is better, along with the frequency accuracy, but the phase lag is somewhat more prevalent, leading to the stiffer test having better phase performance. For frequency response accuracy, as shown in the plotted curves in Figure 9, it is clear that the resonance identification of the DINO is more accurate at lower frequencies, but the overall accuracy of the frequency response curves appears better in higher frequency systems, as illustrated in Table 6.

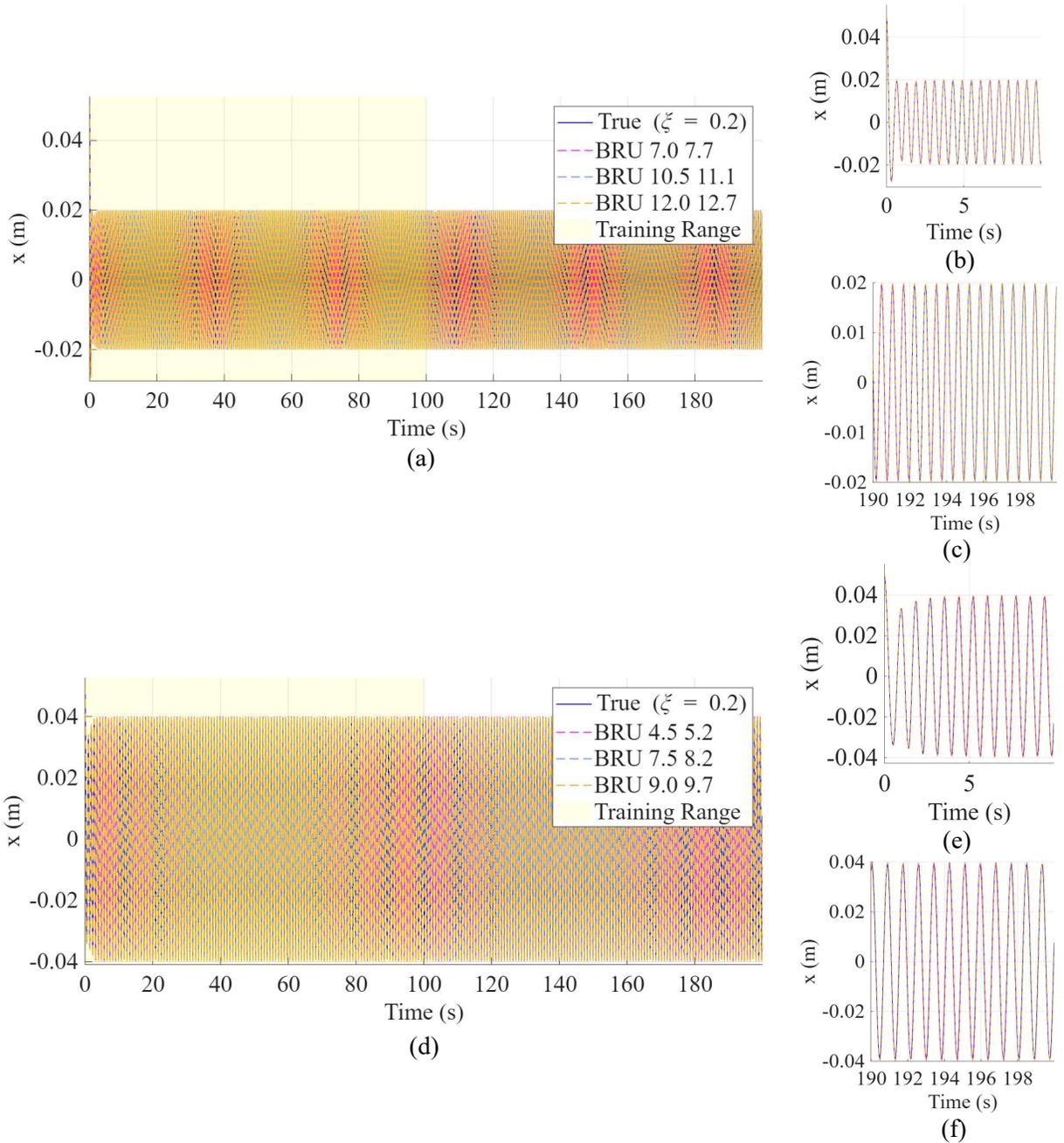

Figure 8. Time response of DINO 3.0 on (a-c) LS-1A and (d-f) LS-1B systems.

Table 5. Summary of time response accuracy of DINO 3.0 on configurations LS-1A and LS-1B.

| DINO 3.0 time response accuracy LS-1A | | | | | |
|---|---|---|---|---|---|
| BRU (rad/s) | MSE (m$^2$) | Mean error (%) | Amp. error (%) | Phase error (%) | Freq. error (%) |
| 0.1-0.8 | 2.56E-06 | 0.07 | 1.92 | 0.05 | 0.30 |
| 0.8-1.5 | 2.99E-06 | 0.08 | 2.09 | 0.05 | 0.30 |
| 1.5-2.2 | 2.91E-06 | 0.08 | 1.87 | 0.05 | 0.30 |
| Accuracy (%) | | 99.92 | 98.04 | 99.95 | 99.70 |

| DINO 3.0 time response accuracy LS-1B | | | | | |
|---|---|---|---|---|---|
| BRU (rad/s) | MSE (m$^2$) | Mean error (%) | Amp. error (%) | Phase error (%) | Freq. error (%) |
| 0.1-0.8 | 7.84E-06 | 0.07 | -0.37 | 0.09 | 0.00 |
| 0.8-1.5 | 8.37E-06 | 0.07 | 0.10 | 0.09 | 0.00 |
| 1.5-2.2 | 8.29E-06 | 0.07 | 0.03 | 0.09 | 0.00 |
| Accuracy (%) | | 99.93 | 99.92 | 99.91 | 100.00 |

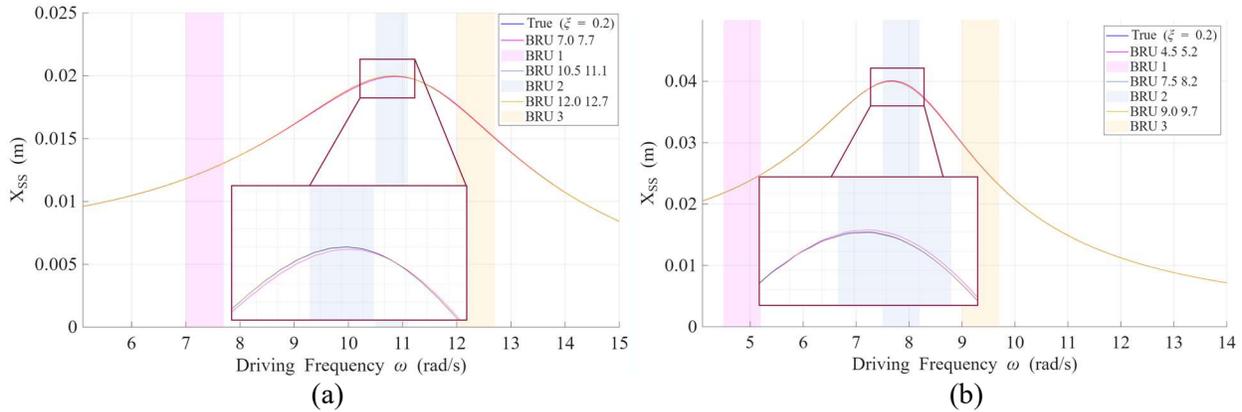

Figure 9. Frequency response curves with a comparison of three training bands: (a) LS-1A and (b) LS-1B.

Table 6. Frequency response accuracy for both dimensional systems.

| DINO 3.0 frequency response accuracy LS-1A | | | | |
|---|---|---|---|---|
| BRU (rad/s) | Shape error (%) | Peak error (%) | Resonance error (%) | $\hat{\omega}_p$ (rad/s) |
| 0.1-0.8 | 0.19 | 0.31 | 0.37 | 10.87 |
| 0.8-1.5 | 0.07 | 0.04 | 0.00 | 10.83 |
| 1.5-2.2 | 0.02 | 0.04 | 0.00 | 10.83 |
| Accuracy (%) | 99.90 | 99.87 | 99.88 | 10.85 |

| DINO 3.0 frequency response accuracy LS1-B | | | | |
|---|---|---|---|---|
| BRU (rad/s) | Shape error (%) | Peak error (%) | Resonance error (%) | $\hat{\omega}_p$ (rad/s) |
| 0.1-0.8 | 0.18 | 0.27 | 0.00 | 7.69 |
| 0.8-1.5 | 0.08 | 0.11 | 0.00 | 7.69 |
| 1.5-2.2 | 0.05 | 0.06 | 0.00 | 7.69 |
| Accuracy (%) | 99.90 | 99.86 | 100.00 | 7.69 |

### 4.3. Base excitation time history and frequency response: DINO 3.0

When writing the equation of motion for a spring-mass-damper under harmonic base excitation, an interesting property is seen as it pertains to the adaptation of a machine learning algorithm. As mentioned in Section 3, the nonhomogeneous forcing term must be a function of the independent variable only (i.e., must only vary with respect to time) for the DINO formulation to work as built. Additionally, the forcing term may not contain any unknown system parameters. This follows naturally from the motivation of the algorithm. If the system parameters were known, a lower-order numerical model could easily be created, so it must be assumed that system's parameters are not available constants. In these problems, it is assumed that the initial condition and forcing function are known at all times. Looking back to equation (13), the base excitation equation of motion has a forcing term that contains the damping and stiffness values, resulting in an invalid form for the DINO. To obtain a tenable form for the system, the data is converted into relative coordinates. This is advantageous because in a relative coordinate frame, the forcing term is independent of the system's physical properties. Using equation (22), the governing equation of motion for the relative displacement is governed by equation (23) as follows:

$$z(t) = x(t) - y(t) \tag{22}$$

$$m(\ddot{x} - \ddot{y}) + c(\dot{x} - \dot{y}) + k(x - y) = mY\omega^2 \cos(\omega t) \tag{23}$$

$$\ddot{z} + 2\xi\omega_n \dot{z} + \omega_n^2 z = Y\omega^2 \cos(\omega t) \tag{24}$$

Because the base excitation is assumed to be known, absolute coordinates may be reclaimed at any time after network inference by addition to the inferred solution $z(t)$. By inspection, both the requirement for conventional forcing and no inclusion of the system's parameters are satisfied by equation (24). For the sake of simplicity, one configuration of ($\omega_n = 1 \frac{rad}{s}$ and $\xi$=0.2) is considered to check the validity of DINO 3.0 for base-excited systems. The results in Figure 10 and Table 7 show a similarity to the results for forced excitation done in Section 4.1, indicating that the relative coordinate learning method is a valid extension of the DINO formulation.

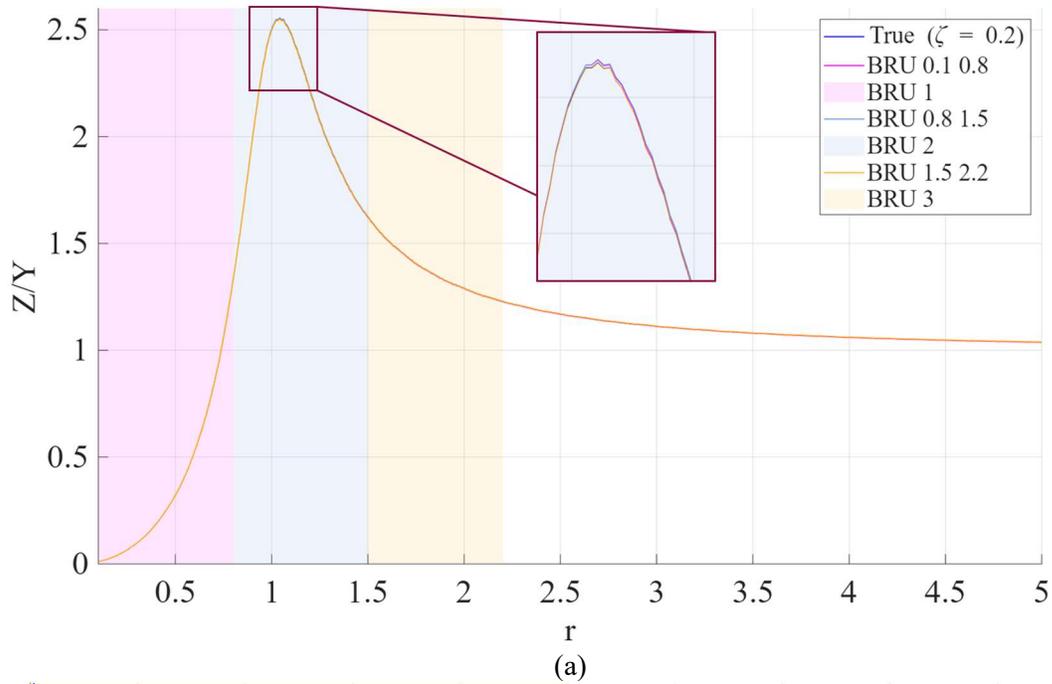

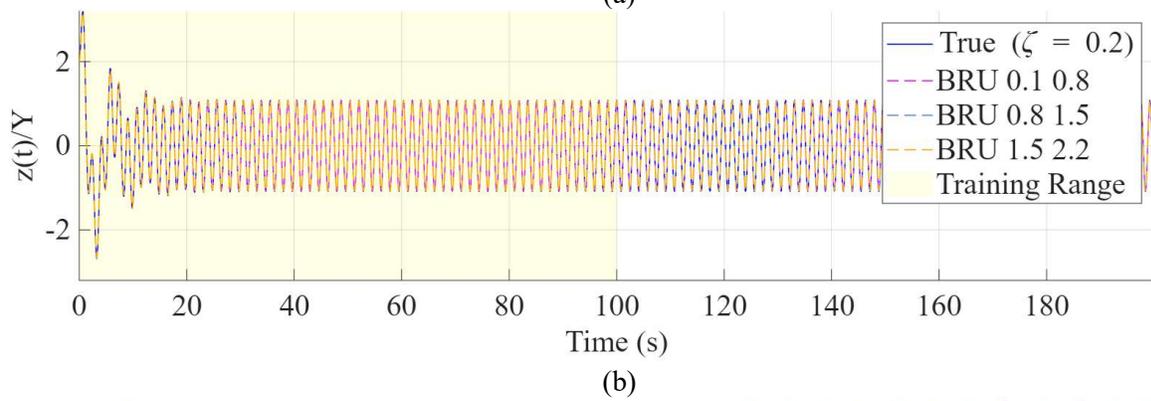

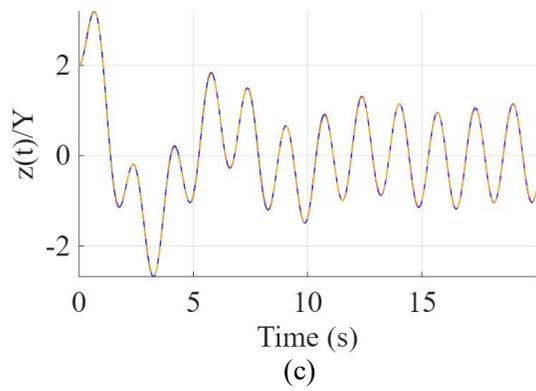 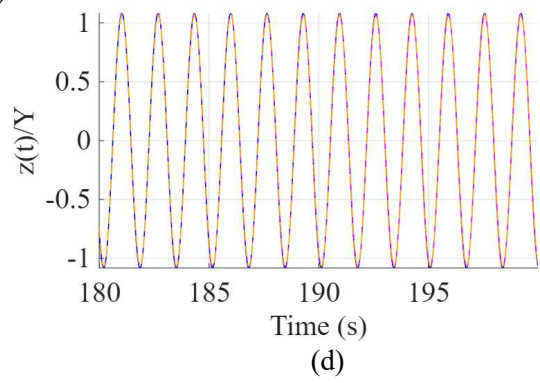

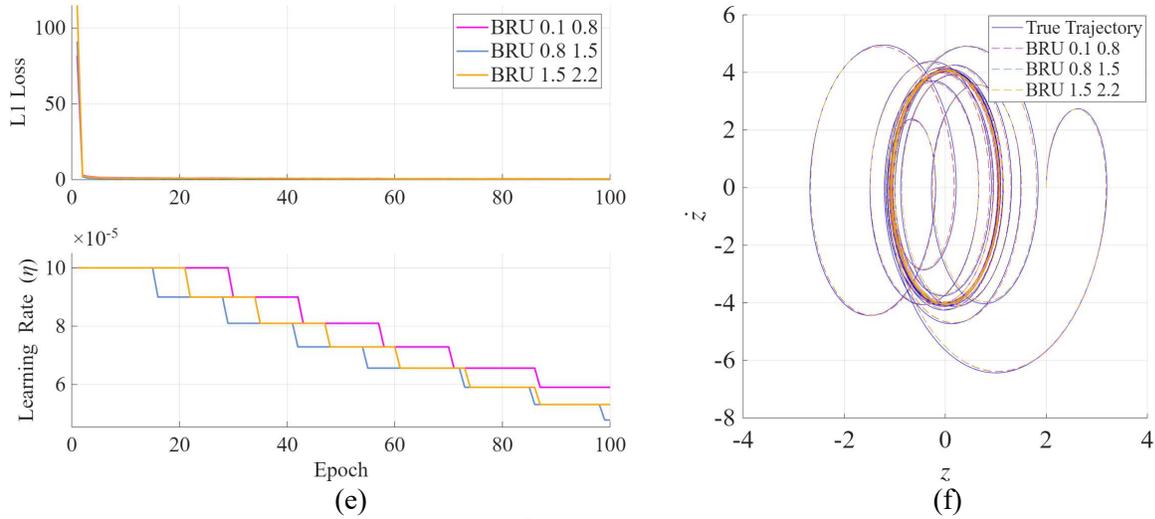

(e)                                      (f)

Figure 10. (a) Frequency response curve of $\frac{Z}{Y}$ on the LS-1 base-excited system using DINO 3.0, (b) time responses of the three test BRU curricula compared to the true time history driven at $r = 3.77$, (c) zoom-in of the transient time condition, (d) zoom-in of the steady-state condition, (e) loss and learning rate procession during 1000 epochs of training, and (f) phase space trajectory of the base-excited system.

Table 7. BRU comparison of DINO 3.0 frequency response accuracy in base excitation inference.

| DINO 3.0 frequency response accuracy (base excitation) | | | | |
|---|---|---|---|---|
| BRU (rad/s) | Shape error (%) | Peak error (%) | Resonance error (%) | $\hat{\omega}_p$ (rad/s) |
| 0.1-0.8 | 6.88 | 0.02 | 0.00 | 0.95 |
| 0.8-1.5 | 6.42 | 0.01 | 0.00 | 0.95 |
| 1.5-2.2 | 6.34 | 0.18 | 0.00 | 0.95 |
| Accuracy (%) | 93.45 | 99.93 | 100.00 | 0.95 |

To further validate the conversion, it is useful to repeat this process for damping ratios between 0.1 and 1.0 to identify the crossover point at $r = \sqrt{2}$ in absolute coordinates. If present, which is the case as indicated in Figure 11, it can be said that DINO 3.0 is valid for base excitation, making it a far more applicable method to experimental setups for beams, plates, and other vibrating systems. The selections of the damping ratios tested have the $r = \sqrt{2}$ crossover point that must exist in base excitation problems. An additional convenience of this formulation is that it may be easily chained to allow the solution of multiple degree-of-freedom systems with no loss of applicability. An interesting additional note to this result is that the overall frequency response curve accuracy increases as the damping is increased. This is congruent with the results obtained in forced excitation. This is likely because the transient solution is more concise and completely contained within the limited training horizon.

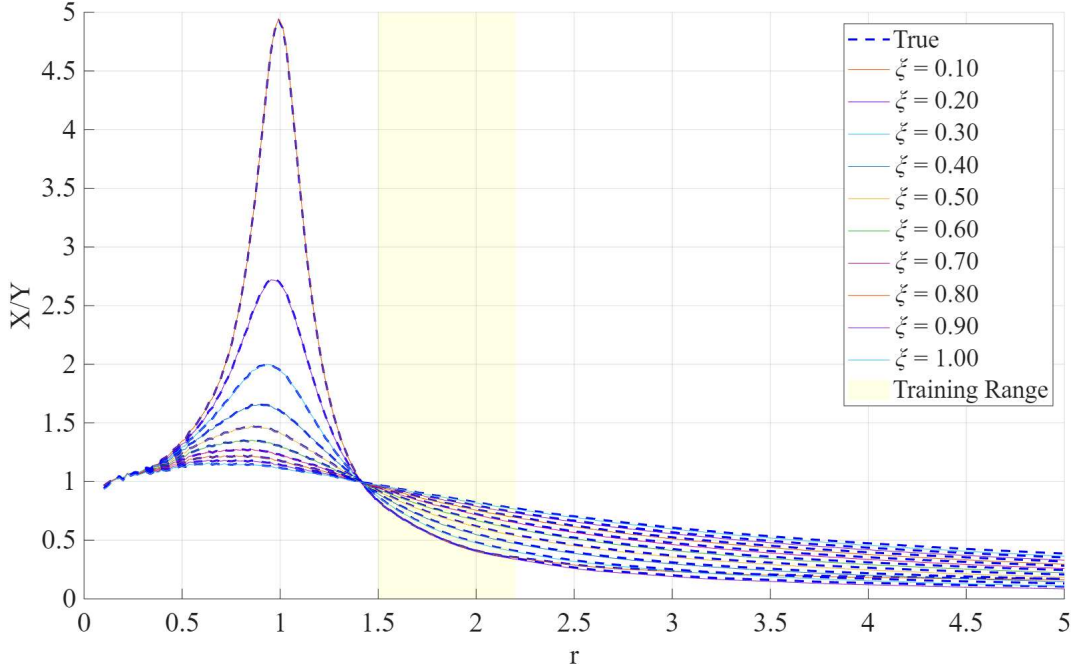

Figure 11. Base excitation results for LS-1 with damping ratios between $\xi = 0.1$ and $\xi = 1.0$.

Table 8. Frequency Response accuracy of DINO 3.0 on the base excitation problem for damping ratios between $\xi = 0.1$ and $\xi = 1.0$.

| | DINO 3.0 frequency response accuracy (base excitation) | | | |
|---|---|---|---|---|
| $\xi$ | Shape error (%) | Peak error (%) | Resonance error (%) | $\hat{\omega}_p$ (rad/s) |
| 0.1 | 0.55 | 0.23 | 0.00 | 0.99 |
| 0.2 | 6.34 | 0.19 | 0.00 | 0.95 |
| 0.3 | 6.14 | 0.18 | 0.00 | 0.93 |
| 0.4 | 6.04 | 0.15 | 0.00 | 0.89 |
| 0.5 | 6.05 | 0.12 | 0.00 | 0.85 |
| 0.6 | 5.72 | 0.13 | 0.00 | 0.85 |
| 0.7 | 5.47 | 0.10 | 0.00 | 0.85 |
| 0.8 | 5.14 | 0.09 | 0.00 | 0.85 |
| 0.9 | 4.89 | 0.02 | 0.00 | 0.64 |
| 1.0 | 4.56 | 0.01 | 0.00 | 0.64 |

## 5. Stability analysis and limitations

All machine learning algorithms are heavily dependent on the properties of training data, and the DINO algorithm is not an exception. In Section 4, the results for several single-degree-of-freedom systems were analyzed to assess the validity of this machine learning formulation on nondimensional, dimensional higher-frequency, and base-excited spring-mass-damper systems. The reader can be relatively confident that, for reasonable, realistic systems, and with the correct training data selection, the DINO can achieve high accuracy on linear systems. How are these BRU bands selected? Are there poor choices that lead to low-fidelity models? How is the stability and accuracy of DINO affected by high frequencies, low sampling rates, and selected hyperparameters? Section 5 aims to answer these questions to gain a more complete picture of how this algorithm interacts with system dynamics and its limitations. In particular, it will be found that better performance is obtained from training bands placed near resonance, with high-amplitude forcing, and when many time histories are selected for training. It will also be shown how the network can be assessed for stability on systems regardless of whether the underlying dynamics are well-characterized. These findings together allow the proper use-case of DINO to be identified.

### 5.1. Training data sensitivity

To fully understand the way that dataset selection affects forecast quality on the frequency response curves in inference, the method used must be analyzed. The training algorithm employed in the above testing is the Banded Random Uniform (BRU) curriculum. In this method, ten trajectories are selected from the normal distribution in a selected frequency band. For consistency in the results, the random uniform generator is set to the same seed for all tests. The selectable parameters here are the bandwidth and location of the BRU to be used for training. To get more general conclusions about which frequency regions are most advantageous to select, a bandwidth variation test and BRU location variation are performed.

Bandwidth variation is a measure of the breadth of frequency data required for the DINO to acquire a sufficiently accurate model of system physics. While the same amount of data (50% of ten forced trajectories) is used for each training trial, the band from which those ten trajectories are chosen is varied. The test is run for 10 trials with BRUs centered at $r = 1.0, 2.0,$ and $5.0$ frequency units for bandwidths varying from 0.1 to 2.0. Because each trial requires a separate instance of training, this test is time-consuming, so it was not performed for the dimensional systems LS-1A and LS-1B, which require higher resolution. Overall, it is found that the bandwidth $\Delta_\omega$ has a weak proportionality to resonant peak error, and the band center location $\omega_T$ is directly proportional to error, as shown in the plotted curves in Figure 12.

Figure 12(a) illustrates the way that the training bandwidth is expanded during this test when $\omega_T = 5.0$. In general, increasing the bandwidth will yield a weak improvement of the DINO peak amplitude accuracy. Figure 12(b) indicates the strength of this improvement when the band center is at the locations $\omega_T = 1.0, 2.0,$ and $5.0$. The change when increasing the bandwidth is weak in terms of peak error, and per Figure 12(c), nonexistent in terms of resonant frequency error. A reasonable widening of the training bandwidth will not improve the identified resonant frequency and will only slightly improve the peak amplitude. The total improvement for choosing the best bandwidth is only 0.8%, which leads to the conclusion that bandwidth is not a critically important hyperparameter.

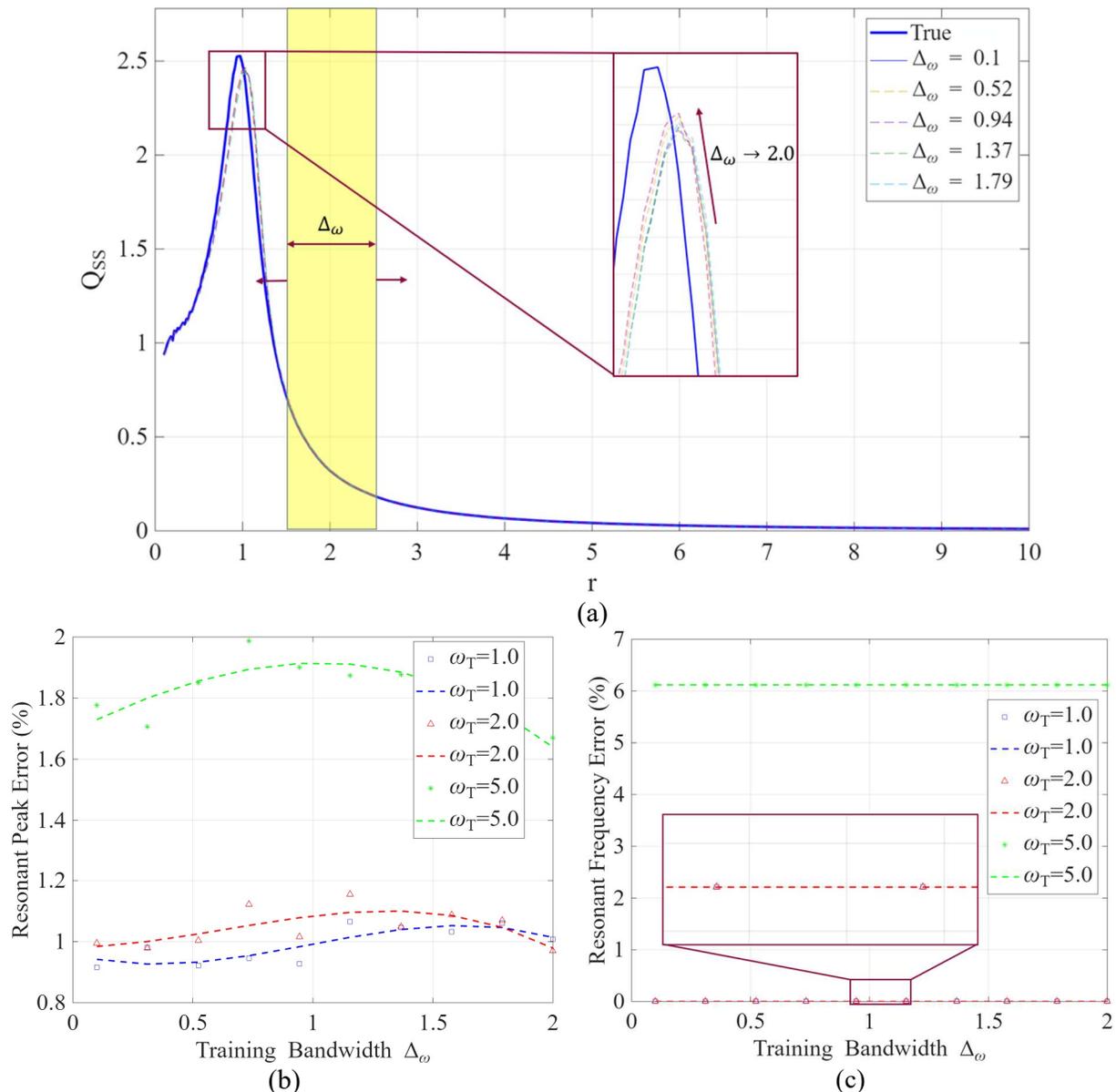

Figure 12. (a) Bandwidth variance for 10 values for $\Delta_\omega \in [0.1, 2.0]$ when $\omega_T = 5$, (b) peak amplitude percentage error versus training bandwidth for three choices of $\omega_T$, and (c) resonant frequency percentage error versus training bandwidth for three choices of $\omega_T$.

Moving on to the importance of the band center location, it is found that there is a much higher sensitivity in Figures 13(b) and 13(c). The plotted frequency response curves in Figure 13(a) indicate that as the training band center is increased, the DINO peak amplitude decreases and the DINO resonant frequency increases. This change can yield up to 7% increases in error when compared to a resonant training center. This effect is expected because the predictions will degrade when the training data is less proximal to the resonance. However, it would be useful to know why this occurs, and whether there are changes to the training process that could mitigate the degradation of DINO performance at training frequencies far from resonance.

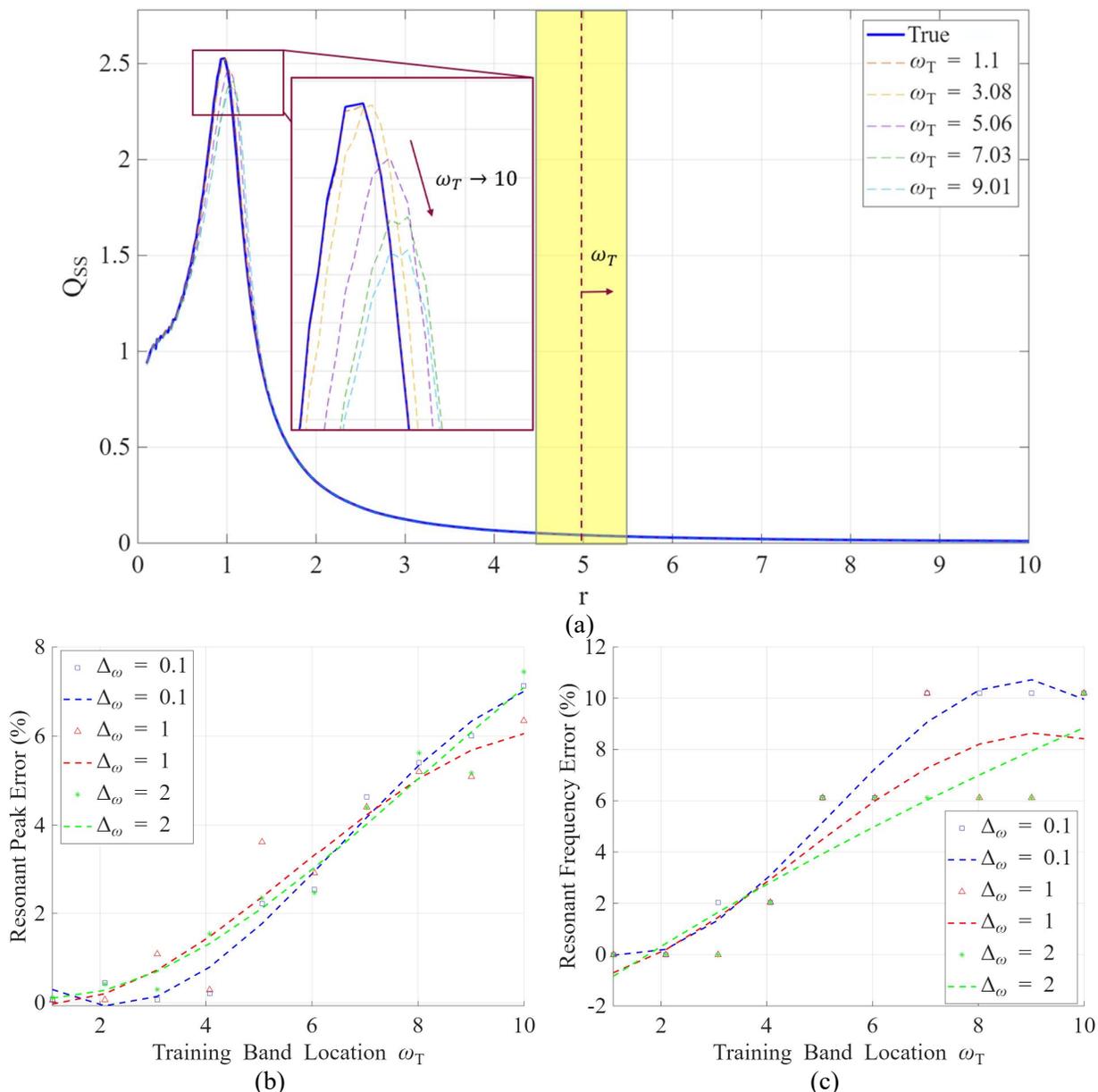

Figure 13. (a) Band center variance for 10 trials $\omega_T \in [1.1, 10]$ when $\Delta_\omega = 0.1$, (b) band center trials for different choices of $\Delta_\omega$ compared via peak percentage error, and (c) resonant frequency error for band center trials.

This proportionality between error and training frequency band center, shown in Figure 13, calls for further analysis. It would indicate that there is less transient content in higher frequencies, but this is simply not true. In general, for linear damped oscillators, the steady-state solution is:

$$x_s(t) = \sqrt{B_1^2 + B_2^2} \cos(\omega t - \phi) \tag{25}$$

where

$$B_1 = \frac{A(\omega_n^2 - \omega^2)}{(\omega_n^2 - \omega^2)^2 + (2\xi\omega_n\omega)^2}, B_2 = \frac{A(2\xi\omega_n\omega)}{(\omega_n^2 - \omega^2)^2 + (2\xi\omega_n\omega)^2}, \phi = \tan^{-1}\left(\frac{2\xi\left(\frac{\omega}{\omega_n}\right)}{1 - \left(\frac{\omega}{\omega_n}\right)^2}\right)$$

Equation (25) states that the steady-state solution only grows smaller with increased driving frequency, but the transient solution remains the same strength throughout, because these frequency response curves are generated with a uniform initial condition. One possible explanation is that the L1 loss function used in training passes larger gradients to the DINO when training on high-amplitude data, causing unusually weak convergence in higher-frequency training. The most natural modification from a machine learning perspective is to perform a batchwise normalization using the root mean square (RMS) amplitude of the training data. It was found in testing of this, however, that it does not significantly improve the $\omega_T$ resonance decay. By revisiting equation (25) again, it could be concluded because $x_s(t)$ grows weaker as $\omega$ increases, the steady-state solution being strong can aid convergence.

The other way to strengthen $x_s(t)$ is to increase the driving amplitude of the training data. To verify this possible reason, the dimensional system must be used for $(\omega_n = 1 \frac{rad}{s}, \xi = 0.2)$. Currently, training is performed for $A = 1 \frac{m}{s^2}$. In Figure 14, the performance of the driving amplitudes of $A = 1.0, 5.0,$ and $10.0 \frac{m}{s^2}$ are compared across the $\omega_T$ spectrum. The $A = 5 \frac{m}{s^2}$ curve of Figure 14 shows a significant moderation of the $\omega_T$ resonance decay phenomenon. Once the training is carried to higher driving amplitudes, the network begins to overpredict the resonant peak. This is an important conclusion in several regards. First, selecting the most effective driving amplitude of the training data can optimize DINO performance without data near resonance. Second, while the DINO is formulated to perform inference with any arbitrary excitation, the choice of training excitation is far from arbitrary, and in fact has significant influence on the results. Because the above results were shown for harmonic excitation training, further investigations in this topic may be needed when considering other types of excitations.

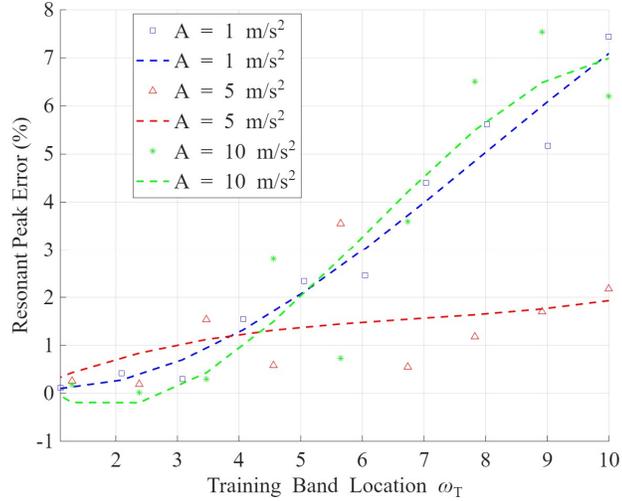

Figure 14. Peak percentage error for $\omega_T$ spectrum for increasing driving amplitudes.

To justify the choice of 10 training trajectories used in all previous results, a convergence of network error with respect to the number of training trajectories is presented in the plotted curves in Figure 15. It becomes apparent from this figure that increasing the number of training trajectories and the length of the trained time history have the completely expected effect of reducing DINO error. Indeed, varying the length of the time history used in training reduces the number of trajectories required to reach a performance plateau, which exists at 10 trajectories for a length of $t_T$=3, 5 trajectories for a length of $t_T$=10, and 3 trajectories for a length of $t_T$ =100. This is a convenient property which indicates that the same dynamical information can be extracted by many short time histories or fewer long time trajectories. It bears mentioning that the gain in trajectory convergence in Figure 15 when increasing the length of a training time history gets smaller, which means that the transient part of the vibration alone contains the most valuable dynamical information. This is not surprising because the transient part of the solution contains the natural frequency and damping of the system.

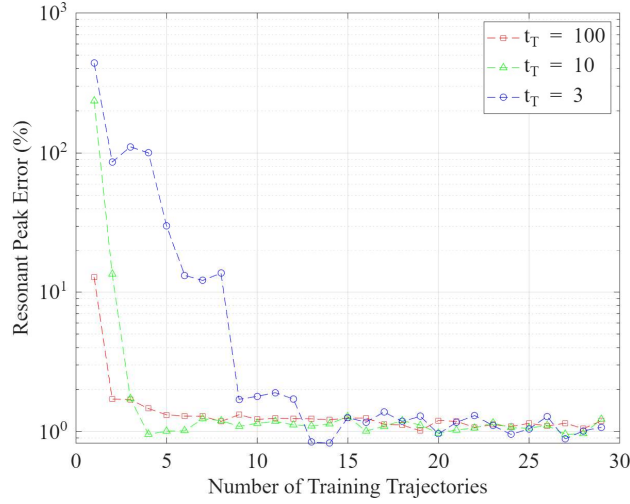

Figure 15. Convergence of network error as the number of training trajectories used is increased for three lengths of transient training time at 100, 10, and 3.

### 5.2. Frequency stability analysis

The tests performed in Section 4.2 provide basic validation for the performance of the DINO when the system stiffness is raised. It was shown that for two realistic choices of peak amplitude (2 cm and 4 cm), the accuracy of the network in time and frequency was essentially identical to lower stiffnesses. A caveat is that as natural frequency is increased, the training sampling frequency must be commensurately increased to satisfy a constant sampling to natural frequency ratio $\frac{f_s}{f_n}$. More generally, it would be an informative exercise to investigate how the accuracy of the network is affected by a continued increase in natural frequency at a constant frequency ratio of $\frac{f_s}{f_n}$.

From prior knowledge of DINO and the systems which underlie its operation, the expectation is that enforcing a uniform frequency ratio will result in below-average errors at low frequencies, and some nonlinear increasing relation between error and driving frequency at high frequency. The average error of the DINO should lie near resonance for any chosen frequency ratio and natural frequency. This suggests that the frequency response curve forecasts degrade when calculated above the specified target frequency, and will become unstable when the driving frequency exceeds a value determined by the sampling to natural frequency ratio.

The local frequency ratio during an inference pass with a sampling frequency $f_s = \frac{1}{h}$, the local frequency ratio can be defined by the local frequency ratio $R_\omega$ found in equation (26). It is expected that the network inference becomes unreliable when the local frequency ratio becomes 2, based on the Nyquist rate, and hence $\omega_{critical} = \frac{\omega_n}{2} R_s$.

$$R_\omega = \frac{2\pi}{\omega h} = \frac{\omega_n}{\omega}\frac{2\pi f_s}{\omega_n} = \frac{\omega_n}{\omega} R_s \qquad (26)$$

For the LS-1 tests previously shown, $R_s$ is approximately 62, so the critical frequency does not occur until 31 times the resonant frequency. To validate the appearance of this critical frequency, DINO tests were run on a dimensional linear oscillator with $A = 1\frac{m}{s^2}$ with $R_s = 10$ on ten systems of increasing natural frequency from 10 to 50 $\frac{rad}{s}$. To ease comparison, the frequency response curves and error responses were normalized in two ways, by plotting using the normalized driving frequency $r = \frac{\omega}{\omega_n}$, and by normalizing amplitude with peak magnitude $X_R$. In the following analysis, the error metric will be the fraction of the mean absolute error of the DINO frequency response curve divided by $X_R$, $E(\%) = 100\% \cdot \frac{E_{MAE}}{X_R}$. This will render the different frequency response curves comparably.

In Figure 16(a), the maximum error occurs in all trajectories near resonance and near the critical frequency limit. For $R_s = 10$, the maximum error at resonance is approximately 1.5% of resonant amplitude. Figure 15(b) is the plot of output displacement divided by the corresponding peak amplitude $\frac{X_{SS}}{X_R}$, and the $r = \frac{R_s}{2}$ critical point is clearly visible. The implication of this data is that while the Nyquist limit is satisfied locally during inference (i.e., $R_\omega > 2$), the DINO response will be stable, but the accuracy is a matter of how much larger than 2 it is. It is instructive to compare the same FRF for different frequency ratios as shown in Figure 16(c). In Figure 16(d), the percentage error metric is plotted against the frequency ratio $R_s$, and shows the expected inverse proportionality. The DINO inference error is always a combination of network approximation error and numerical truncation error (with an additional float32 roundoff error contributed by both systems). The convenience of this formulation is that increasing $R_s$ can reduce both approximation and truncation error magnitudes, allowing a greater overall convergence.

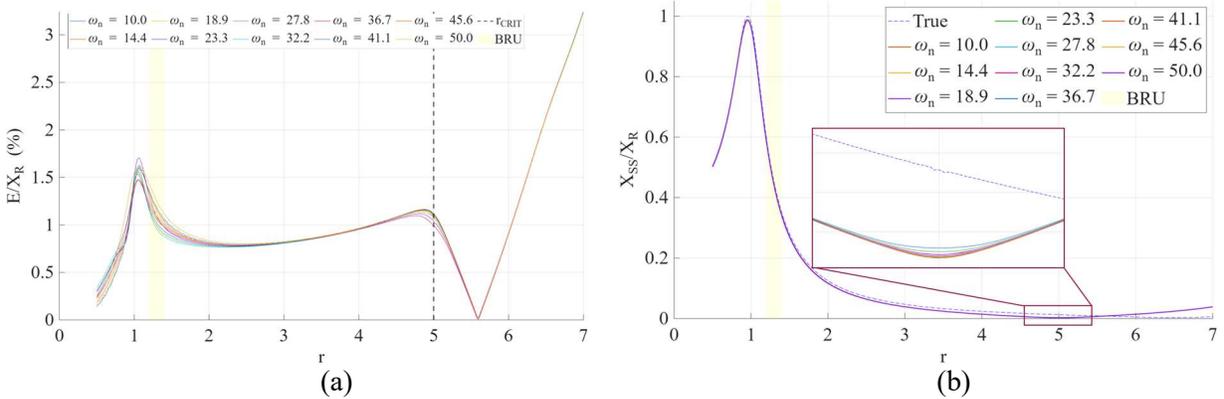

(a)    (b)

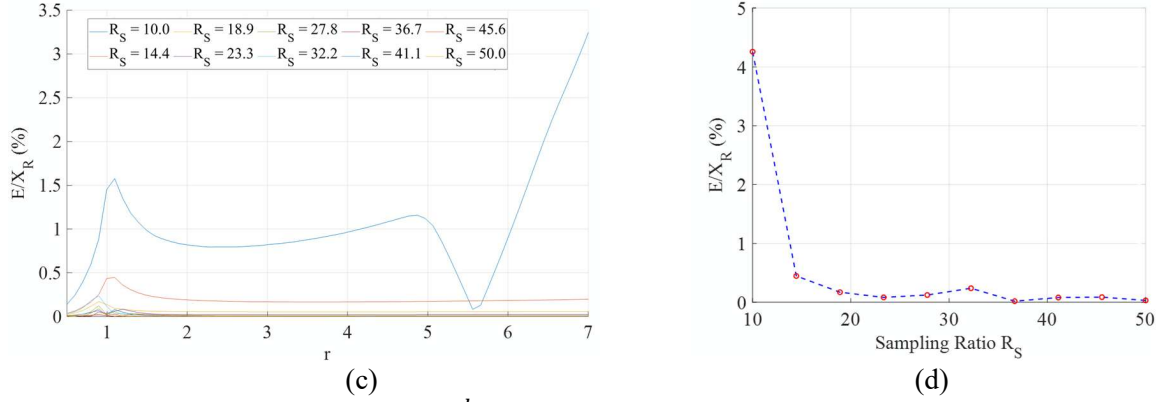

(c)    (d)

Figure 16. (a) Error response for $\omega_n(\frac{rad}{s}) = [10, 50]$ run until $r = 7$, (b) normalized frequency responses for $\omega_n(\frac{rad}{s}) = [10, 50]$ run to $r = 7$, (c) error responses for frequency ratios $R_S = [10, 50]$, $\omega_n = 10\frac{rad}{s}$, and (d) peak error compared to sampling ratio.

### 5.3. Network stability analysis

The question of network stability is essential to determine whether the identified system shares equilibria and dissipative behavior with the training data, and the more global system data. Additionally, it is important to consider the presence of regions in the manifold of $\tilde{G}(\vec{x})$ that are nonphysical and would result in overall instability upon forecasting. A useful philosophy is to see how network stability evolves epoch by epoch, as in [44]. By linearizing the system and studying its free response in the neighborhood of its equilibria, the network's identified equilibrium can be analyzed for stability. Any search of a trained network space will provide limited guarantees of network stability. It is entirely possible for the network to diverge at high amplitudes or allow unstable responses, and this method of stability verification is very much a local check for consistency. Global guarantees of network stability are a different matter in machine learning and will not be discussed here.

For the nondimensional LS-1 system, the equilibrium point lies at $(0, 0)$ in phase space, making this a very straightforward linear stability analysis problem as follows:

$$f(\vec{x}) = \begin{bmatrix} \dot{x}_1 \\ \dot{x}_2 \end{bmatrix} = \begin{bmatrix} x_2 \\ -2\xi x_2 - x_1 \end{bmatrix} \tag{26}$$

$$\frac{\partial f(\vec{x})}{\partial \vec{x}}\bigg|_{\vec{x}=0} = A = \begin{bmatrix} 0 & 1 \\ -1 & -2\xi \end{bmatrix} \tag{27}$$

The characteristic equation may be identified as:

$$\det(A - I\lambda) = \lambda^2 + 2\xi\lambda + 1 = 0 \tag{28}$$

$$\lambda = -\xi \pm \sqrt{\xi^2 - 1} = -0.2 \pm 0.9798j$$

Because both real parts of the eigenvalues are negative, this equilibrium point is fully stable. The next step is to test the trained DINO network to see if the same condition is satisfied. It is not necessary that the DINO solution be more dissipative than the training data (this is a matter of accuracy, not stability), but it is fully required that $Re(\tilde{\lambda}_1)$ and $Re(\tilde{\lambda}_2)$ be less than zero. The test process is simple to apply because the Jacobian of the network can be obtained directly at any time using automatic differentiation. Using DINO 3.0 on LS-1, it attains eigenvalues of $\tilde{\lambda} = -0.2 \pm 0.9795j$. From this, it can be quickly found that the network inference will be slightly under-frequency after training, with nearly identical damping to the original system. Linear stability analysis of a known equilibrium is a computationally inexpensive diagnostic to check the network for prediction before inference is performed. By leveraging this test at every training epoch, a new stability diagnostic can be constructed by plotting intermediate network zeros on a real-imaginary plane.

Figures 17(a) and 17(b) indicate that while the DINO does converge on a high-accuracy solution, it reaches a point before epoch 100 where it seeks around the exact solution. This is common in neural networks and visualizing this convergence in the real-imaginary creates a means of network interpretability. It allows for a real-time estimation of the model convergence. Because the real component measures dissipation and the imaginary component measures frequency, it can be deduced that damping uncertainty is comparable to frequency uncertainty for learning rates near $10^{-5}$. Another characteristic is that the network has an initial location in the root-locus plot of $\tilde{\lambda}^0 = -0.0033 \pm 0.0121j$, very near the neutral point, a condition governed by the random number seed used to generate the initial weights. Summarizing these findings leads to the conclusion that real-imaginary visualization of training can allow better hyperparameter initialization and render a more convergent result in fewer epochs.

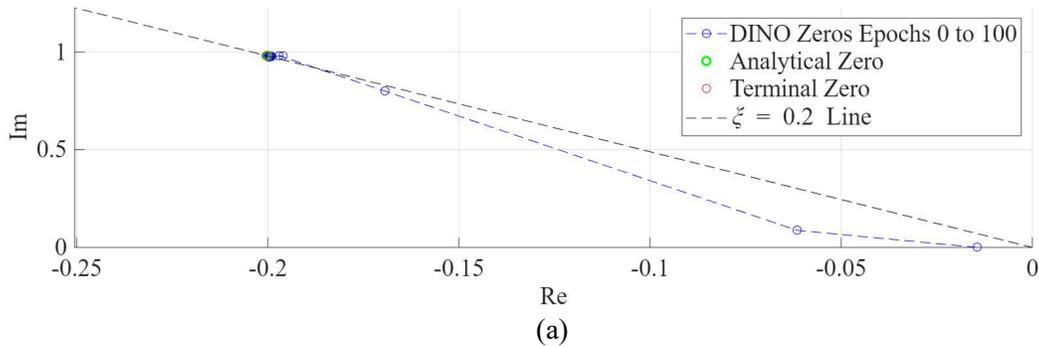

(a)

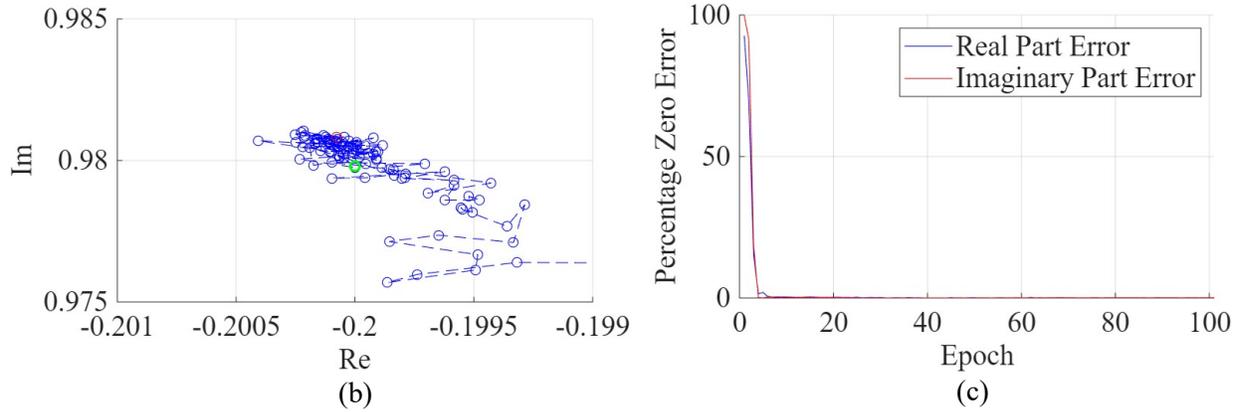

Figure 17. (a) Eigenvalue procession of operator zeros from epochs 0 to 100, (b) searching phase of model eigenvalues, and (c) percentage error of real and imaginary components. Constant $\xi = 0.2$ line denoted in black.

During a conventional run of the DINO on LS-1, the convergence lies slightly below the constant $\xi$ line denoted in black on Figure 17(a). These processions indicate how the training approaches a final solution, with the damping identified early, and frequency gradually increased as the epochs continue, eventually reaching the final solution. Further testing using this diagnostic can also verify that in the case of an unlucky randomly generated network initial condition, where $\tilde{\lambda}^0$ is right of the marginal line, that training will still proceed to stability. The DINO formulation inherits the convenience of stability analysis from Neural ODEs. It is known that dynamical systems must have a divergence in phase space that is less than or equal to zero unless an external force is applied. If this condition is not satisfied by network predictions, the result can be called nonphysical [45].

6. **Observations and discussion**

Version progression demonstrated how architectural refinement improved both convergence and accuracy. DINO 1.0 exhibited amplitude and phase errors of 13.92% and 9.32%, respectively, with errors caused primarily by phase lag and bias. In contrast, DINO 3.0 realized a 99.99% accurate time response and 99.90% frequency response curves accuracy across all BRU training regions, with shape, peak, and resonance errors each below 0.3%. Development of the architecture and adoption of the Hilbert latent space demonstrate improved spectral alignment of DINO 3.0, as shown qualitatively in Figure 6 for each version. It was shown that computational efficiency also increased, with training time decreased by approximately 13% relative to earlier models. The reduction was likely due to the elimination of time encoding and explicit forcing terms in the input tensor, enabling faster training on a smaller dataset. Dimensional LS-1 testing confirmed that DINO's extrapolation ability extends to physical systems with higher natural frequencies. The 2 cm and 4 cm LS-1 cases retained >99.90% mean and frequency accuracy, with minor phase lag under higher stiffness. Frequency response curve predictions improve in higher-frequency systems even when the sampling ratio is fixed, which clarifies the action of the numerical method used in reducing error. This suggests that DINO generalizes not only to unseen periods in a time history, but also to unseen frequencies and time resolutions.

The stability analysis established that forecast degradation follows a predictable pattern as the local frequency ratio approaches the Nyquist limit, retaining stability up to the predicted critical frequency limit at $\omega_{critical} = \frac{\omega_n}{2} R_s$. The algorithm indicated an ability on linear systems to perform accurate forecasting using 7% of the dataset frequency domain, training on 50% of the provided time histories. The tests performed here, in addition to serving as basic validation and stability diagnostics, also suggest promising insights for future experimentation. Tests show that during BRU comparisons, training data in post-resonance yielded higher fidelity forecasts than pre-resonant training. When testing components in vibration analysis, the main concern is locating the frequencies and amplitudes of resonance, so experimental tests using DINO extrapolation should be trained with mid-range frequency data in an experimental set. This allows extrapolation to be performed into lower frequency regions, where the resonances can be accurately identified. The training dataset should be carefully normalized batchwise to ensure more uniform performance between training bands. The tests also demonstrated that the most reliable training data for the algorithm lies in early post-resonance for linear single degree-of-freedom systems. Additionally, as expected, the model error decreases slowly as the bandwidth of the BRU training band is increased.

This algorithm may be applied in the future to vibrating systems with nonlinear stiffness, nonlinear damping, multiple degrees of freedom, and structural complexities like bolted joints and contacts. The key idea is the construction of a unified methodology that allows precise models to be created with more general applicability. In particular, bolted joint analysis can benefit from a more consistent data-driven modeling method [46]. These systems contain complex time-dependent characteristics and may be composed and analyzed from knowledge of the frequency response data of its components by sub-structuring [47, 48]. These studies proposed the use of nonlinear models that capture different types of friction in the joint. These models will serve as a preliminary data source for training, as an experimentally fed model is produced to tolerate the vast complexities found in real systems. This model may, through the methods established above, allow better comprehension of bolted-joint behaviors in vibration.

From a machine learning and numerical analysis perspective, there is also much more to be learned about the stability, training convergence, and initialization of the DINO, especially when analyzing how different initialization strategies in the literature [43] affect the stability and initial condition of the network at epoch zero of a training run. While this, and other tactics like using multistep methods in the integration [24], adapting the system for stiffness [7], and the use of more advanced optimizers lie in the realm of algorithm optimization, they still stand as important future steps to expand the algorithm to perform on systems of higher complexity.

## 7. Conclusions

In modern experimental modal analysis of engineered parts, there can be a large time requirement, expense, and difficulty in accurately determining a system's FRF. Whether due to time constraints, equipment limitations, or other difficulties in characterization, some regions of the FRF were more convenient to obtain experimentally than others. Machine learning algorithms have achieved great advancements in accuracy and extrapolation fidelity on dynamical problems, and the fusion of the DeepONet and Neural ODEs has been demonstrated in this investigation to be an advantageous combination for the solution of vibration problems. The methodology behind the DINO fused and modified recent developments in machine learning to perform extrapolation on limited vibrating system data to elicit an accurate frequency response curve. This forecast numerically solves the target data of an experiment, using only dynamics from the system of inquiry. The use of more complete numerical models will allow faster design iteration in the creation of aerospace components. Faster design iteration can lead to more mature prototypes, in turn allowing for more extensive field testing. The acceleration afforded by machine learning in this process makes a workflow that can help prevent unforeseen vibration failures, a key objective in vibration analysis.

**Acknowledgments**

The authors D. Bluedorn and A. Abdelkefi would like to acknowledge financial support from New Mexico Consortium and Sandia National Laboratories (SNL). Sandia National Laboratories is a multimission laboratory managed and operated by National Technology \& Engineering Solutions of Sandia, LLC, a wholly owned subsidiary of Honeywell International Inc., for the U.S. Department of Energy's National Nuclear Security Administration under contract DE-NA0003525. This paper describes objective technical results and analysis. Any subjective views or opinions that might be expressed in the paper do not necessarily represent the views of the U.S. Department of Energy or the United States Government.